\documentclass{IEEEtran4PSCC}
\ifCLASSINFOpdf
   \usepackage[pdftex]{graphicx}
\else
   \usepackage[dvips]{graphicx}
\fi
\usepackage{multirow}
\usepackage[table,xcdraw]{xcolor}
\usepackage{cite}
\usepackage{booktabs}

\usepackage[cmex10]{amsmath}
\usepackage{amsfonts}

\DeclareMathOperator*{\argmin}{arg\,min}
\usepackage{multirow}

\usepackage{subcaption}

\usepackage{url}

\makeatletter
\let\old@ps@headings\ps@headings
\let\old@ps@IEEEtitlepagestyle\ps@IEEEtitlepagestyle
\def\psccfooter#1{%
    \def\ps@headings{%
        \old@ps@headings%
        \def\@oddfoot{\strut\hfill#1\hfill\strut}%
        \def\@evenfoot{\strut\hfill#1\hfill\strut}%
    }%
    \def\ps@IEEEtitlepagestyle{%
        \old@ps@IEEEtitlepagestyle%
        \def\@oddfoot{\strut\hfill#1\hfill\strut}%
        \def\@evenfoot{\strut\hfill#1\hfill\strut}%
    }%
    \ps@headings%
}
\makeatother

\begin{document}
%
\title{Learning Optimization Proxies for Large-Scale Security-Constrained Economic Dispatch}

\author{
\IEEEauthorblockN{Wenbo Chen, Seonho Park, Mathieu Tanneau, Pascal Van Hentenryck}
\IEEEauthorblockA{Georgia Institute of Technology\\
\{wenbo.chen, spark719, mathieu.tanneau\}@gatech.edu, pascal.vanhentenryck@isye.gatech.edu}
}


\maketitle

\begin{abstract}
The Security-Constrained Economic Dispatch (SCED) is a fundamental optimization model for Transmission System Operators (TSO) to clear real-time energy markets while ensuring reliable operations of power grids.
In a context of growing operational uncertainty, due to increased penetration of renewable generators and distributed energy resources, operators must continuously monitor risk in real-time, i.e., they must quickly assess the system's behavior under various changes in load and renewable production.
Unfortunately, systematically solving an optimization problem for each such scenario is not practical given the tight constraints of real-time operations.
To overcome this limitation, this paper proposes to learn an optimization proxy for SCED, i.e., a Machine Learning (ML) model that can predict an optimal solution for SCED in milliseconds.
Motivated by a principled analysis of the market-clearing optimizations of MISO, the paper proposes a novel ML pipeline that addresses the main challenges of learning SCED solutions, i.e., the variability in load, renewable output and production costs, as well as the combinatorial structure of commitment decisions.
A novel Classification-Then-Regression architecture is also proposed, to further capture the behavior of SCED solutions.
Numerical experiments are reported on the French transmission system, and demonstrate the approach's ability to produce, within a time frame that is compatible with real-time operations, accurate optimization proxies that produce relative errors below $0.6\%$.
\end{abstract}

\section{Introduction}
\label{sec:intro}

The \textit{Security-Constrained Economic Dispatch} (SCED) is a fundamental optimization model for Transmission System Operators (TSOs) to clear real-time energy markets while ensuring reliable operations of power grids \cite{conejo2018power}.
In the US, TSOs like MISO and PJM execute a SCED every five minutes, which means that the optimization problem must be solved in an even tighter time frame, i.e., well under a minute \cite{Chen2018_MarketClearingSoftware}.
Security constraints, which enforce robustness against the loss of any individual component, render SCED models particularly challenging for large systems \cite{Chiang2015_SolvingSCOPF,Wang2016_SolvingCorrectiveSCOPF,velloso2021exact} unless only a subset of contingencies is considered.
With more distributed resources and increased operational uncertainty, such computational bottlenecks will only become more critical \cite{Chen2018_MarketClearingSoftware}.

This paper is motivated by the growing share of renewable generation, especially wind, in the MISO system, which calls for risk-aware market-clearing algorithms.
One particular challenge is the desire to perform risk analysis in real time, by solving a large number of scenarios for load and renewable production \cite{werho2021_ScenarioGenerationWind}.
However, systematically solving many SCED instances is not practical given the tight constraints of real-time operations. To overcome this computational challenge, this paper proposes to learn an optimization proxy for SCED, i.e., a Machine Learning (ML) model that can predict an optimal solution for SCED, within acceptable numerical tolerances and in milliseconds.

It is important to emphasize that the present goal is not to replace optimization-based market-clearing tools.
Instead, the proposed optimization proxy provides operators with an additional tool for risk assessment.
This allows to quickly evaluate how the system would behave under various scenarios, without the need to run costly optimizations. In particular, because these predictions are combined into aggregated risk metrics, small prediction errors on individual instances are acceptable.

\subsection{Related Literature and Challenges}
\label{sec:intro:motivation}

    The combination of ML and optimization for power systems has attracted increased attention in recent years. A first thread \cite{pan2019deepopf, fioretto2020predicting, chatzos2020high, lei2020data, chatzos2021spatial, velloso2020combining, zamzam2020learning, owerko2020optimal} uses ML models to predict an optimal solution to the Optimal Power Flow (OPF) problem.
    Pan et al. \cite{pan2019deepopf} train a Deep Neural Network (DNN) model to predict solutions to the security-constrained DC-OPFs, and report results on systems with no more than 300 buses. One common limitation of ML models, also noted in \cite{pan2019deepopf}, is that predictions are not guaranteed to satisfy the original problem's constraints. To address this limitation, recent efforts \cite{fioretto2020predicting, chatzos2020high} integrate Lagrangian duality into the training of DNNs in order to capture the physical and operational constraints of AC-OPFs. A similar approach is followed in Velloso et al. \cite{velloso2020combining} in the context of preventive security-constrained DC-OPF. Namely, a DNN is trained using Lagrangian duality techniques, then used as a proxy for the time-consuming master problem in a column-and-constraint generation algorithm. More recently, Chatzos et al. \cite{chatzos2021spatial} embed the Lagrangian duality framework in a two-stage learning that exploits a regional decomposition of the power network, enabling a more efficient distributed training.
    
    Another research thread is the integration of ML models within optimization algorithms, in order to improve runtime performance. For instance, a number of papers (e.g., \cite{deka2019learning,misra2021learning,xavier2021learning, yang2020fast,guha2019machine}) try to identify active constraints in order to reduce the problem complexity. In \cite{xavier2021learning}, the authors investigate learning feasible solutions to the unit commitment problem. Venzka el al. \cite{venzke2020neural} use a DNN model to learn the feasible region of a dynamic SCOPF and transform the resulting DNN into a mixed-inter linear program.
    
    Existing research papers in ML for power systems typically suffer from two limitations. On the one hand, most papers report numerical results on small academic test systems, which are one to two orders of magnitude smaller than real transmission systems. This is especially concerning, as higher-dimensional data has an adverse impact on convergence and accuracy of machine-learning algorithms.
    On the other hand, almost all papers rely on artificially-generated data whose distribution does not capture the variability found in actual operations. For instance, only changes in load are considered, typically without capturing spatio-temporal correlations. Other sources of uncertainty, such as renewable production, are not considered, nor is the variability of economic bids.
    Finally, changes in commitment decisions throughout the day are rarely addressed, although they introduce non-trivial, combinatorial, distribution shifts.

\subsection{Contributions and Outline}
\label{sec:intro:contributions}

    To address the above limitations, the paper proposes a novel ML pipeline that is grounded in the structure of real-world market operations. The approach leverages the TSO's forward knowledge of 1) commitment decisions and 2) day-ahead forecasts for load and renewable productions. Indeed, both are available by the time the day-ahead market has been executed. Furthermore, the paper presents an in-depth analysis of the real-time market behavior, which informs a novel ML architecture to better capture the nature of actual operations.
    
    \begin{figure}[!t]
        \centering
        \includegraphics[width=0.95\columnwidth]{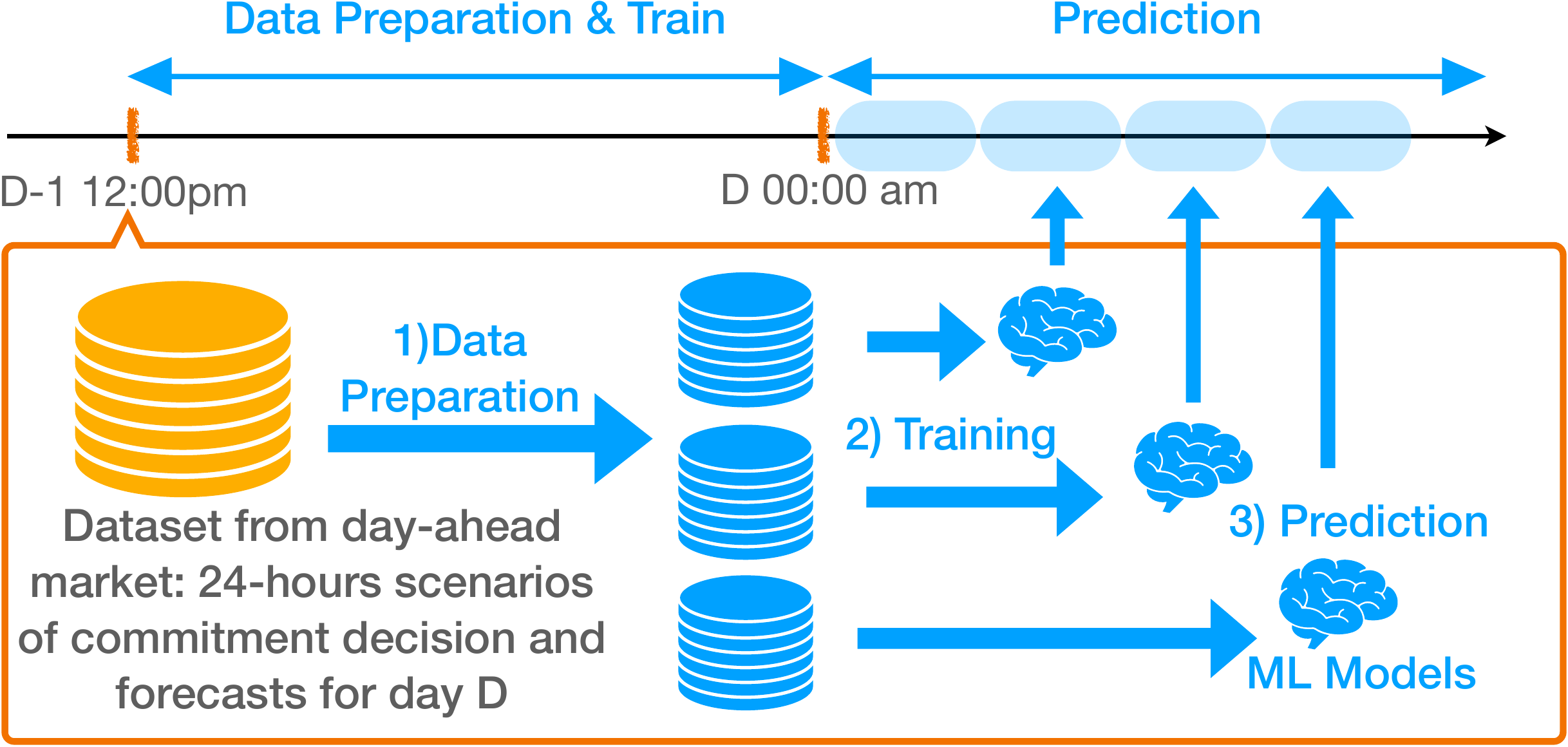}
        \caption{The proposed machine learning pipeline.}
        \label{fig:learning_pipeline}
    \end{figure}
    
    The proposed learning pipeline is depicted in Figure~\ref{fig:learning_pipeline}. First, commitment decisions and day-ahead forecasts for load and renewable generation are gathered, following the clearing of the day-ahead market. Second, this information is used to generate a training dataset of SCED instances, using classical data augmentation techniques. Third, specialized ML models are trained, ideally one for each hour of the operating day, thereby alleviating the combinatorial explosion of commitment decisions: each ML model only needs to consider one set of commitments and focus on load/renewable scenarios around the forecasts for that hour. Fourth, throughout the operating day, the trained models are used in real time to evaluate a large number of scenarios. The entire data-generation and training procedure is completed in a few hours.

    The rest of the paper is organized as follows. Section \ref{sec:opt_pipeline} describes the interplay between day-ahead and real-time markets in the MISO system, and gives an overview of the real-time SCED for MISO. Section \ref{sec:methodology} analyzes the behavior of the real-time market solutions, proposes a combined Classification-Then-Regression architecture, and presents the overall ML pipeline. Numerical experiments on a real-life system are reported in Section \ref{sec:results}

\section{Overview of MISO's Market-Clearing Pipeline} 
\label{sec:opt_pipeline}

This section describes the interplay between MISO's day-ahead and real-time markets; the reader is referred to \cite{BPM_002} for a detailed overview of MISO's energy and reserve markets.

\subsection{MISO Optimization Pipeline}

The day-ahead market consists of two phases, and is executed every day
at 10am.  First, a day-ahead security-constrained unit commitment
(DA-SCUC) is executed: it outputs the commitment and regulation status
of each generator for every hour of the following day.  Then, a
day-ahead SCED is executed to compute day-ahead prices and settle the
market. The results of the day-ahead market clearing are then posted
online at approximately 1pm, i.e., there is a delay of several hours
before the commitment decisions take effect.  While MISO may commit
additional units during the operating day, through out-of-market
reliability studies, in practice, $99\%$ of commitment decisions are
decided in the day-ahead market
\cite{Chen2018_MarketClearingSoftware}.  Accordingly, for simplicity
and without loss of generality, this paper assumes that commitment
decisions from the DA-SCUC are not modified afterwards.

Then, throughout the operating day, the real-time market is executed
every 5 minutes.  This real-time SCED (RT-SCED) adjusts the dispatch
of every generator in response to variations in load and renewable
production, and maximizes economic benefit.  Despite its short-time
horizon, the RT-SCED must still account for uncertainty in load and
renewable production. In the MISO system, this uncertainty mainly
stems from the intermittency of wind farms, and from increasing
variability in load.  The latter is caused, in part, by the growing
number of behind-the-meter distributed energy resources (DERs) such as
residential storage and rooftop solar.  Therefore, operators
continuously monitor the state of the power grid, and may take
preventive and/or corrective actions to ensure safe and reliable
operations.

\subsection{RT-SCED Formulation}

The RT-SCED used by MISO is a DC-based linear programming formulation that co-optimizes energy and reserves \cite{MISO2009_SCED}.
The reader is referred to \cite{BPM002_D} for the full mathematical formulation; only its core elements are described here.
The computation of market-clearing prices is beyond the scope of this paper, and is therefore not discussed here.

The RT-SCED model comprises, for each generator, one variable for energy dispatch and up to four categories of reserves: regulating, spinning, online supplemental, and offline supplemental.
In addition, each generator is subject to ramping constraints and individual limits on energy and reserve dispatch.
Energy production costs are modeled as piece-wise linear convex functions.
Each market participant submits its own production costs and reserve prices via MISO's market portal, and may submit different offers for each hour of the day.
For intermittent generators such as solar and wind, the latest forecast is used in lieu of binding offers.

At the system level, line losses are estimated in real-time from
MISO's state estimator, and incorporated in the formulation using a
loss factor approach as in \cite{FERC2017_MarginalLossCalculation}.
Transmission constraints are modeled using PTDF matrices, where flow
sensitivities with respect to power injections and withdrawals are
provided by an external tool.  Reserves are dispatched on individual
generators, in order to meet zonal and market-wide minimum
requirements. Additional constraints ensure that, in a contingency
event, reserves may be deployed without tripping transmission lines.
Power balance, reserve requirements, and transmission limits are soft,
i.e., they may be violated, albeit at a reasonably high cost.

In summary, the RT-SCED receives the following inputs: the commitment decisions from DA-SCUC, the most recent forecast for load and renewable production, the economic limits and production costs of the generators, the current state estimation, and the transmission constraints and reserve requirements. The RT-SCED produces as outputs the active power and reserve dispatch for each generator.

\section{The Learning Methodology}
\label{sec:methodology}

This section reviews the learning methodology for the RT-SCED.  First,
Section \ref{sec:pattern_dispatch} presents an analysis of the
behavior of optimal SCED solutions in MISO's optimization pipeline.
These patterns motivate a novel ML architecture, which is described in
Section \ref{sec:ml_model}, followed by an overview of the proposed ML
pipeline in Section \ref{sec:ML_pipeline}.  Further details on the
data are given in Section \ref{sec:results}, as wells as in
\cite{PSCC2022-data}.

\subsection{Pattern Analysis of Optimal SCED solutions} 
\label{sec:pattern_dispatch}
    \begin{figure}[!t]
        \centering
        \includegraphics[width=0.95\columnwidth]{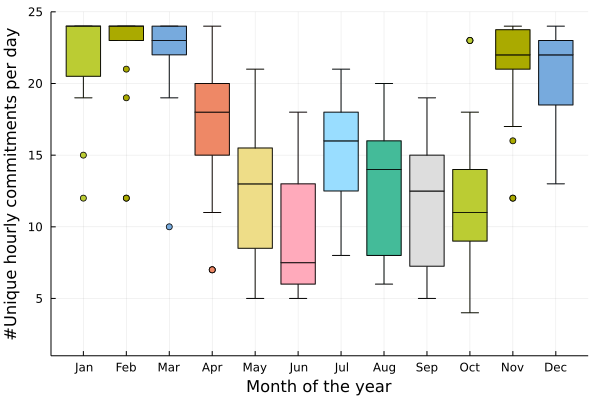}
        \vspace{0.2cm}
        \caption{Distribution of unique hourly commitments per day, for each month of the year 2018. Higher values indicate higher variability in commitment decisions.}
        \label{fig:daily_commitment}
    \end{figure}
    
    The system at hand is the French transmission system, whose network topology is provided by the French TSO, RTE.
    It contains $6{,}708$ buses, $8{,}965$ transmission lines and transformers, and $1{,}890$ individual generators, $1{,}177$ of which are wind and solar generators and $713$ are conventional units.
    The MISO optimization pipeline, described in Section \ref{sec:opt_pipeline}, is replicated on this system for the entire year 2018, yielding $365$ DA-SCUC instances and about $100{,}000$ RT-SCED instances.
    Relevant statistics are reported next.
    
First, unsurprisingly, commitment decisions display a high variability; for example, in 2018, a total of $5{,}380$ different hourly commitments were recorded across the total $8{,}760$ hours of the year, where each ``hourly commitment" is a binary vector of size
713 that contains the (hourly) commitment status of conventional generators. 
The intra-day variability of commitment decisions follows a seasonal pattern, which is illustrated in Figure \ref{fig:daily_commitment}. 
Namely, for each month of the year, Figure \ref{fig:daily_commitment} displays the distribution, across every day of the month, of the number of unique hourly commitments over a day.  The higher values correspond to the higher variability in commitment decisions, while the lower values indicate that the commitment decisions are stable throughout the day.  Typically, the variability of commitment decisions is lower in summer and higher in winter; this behavior is expected since more generators are online in winter, which naturally tends to yield more diverse commitments.
Indeed, in June 2018, all the days have at least 5 and at most 19 different hourly commitments, with $50\%$ of days having less than 7 and $50\%$ having more than 8.  In contrast, in January 2018, except for two outliers, every day has at least 19 different commitments, and 16 days had a different commitment decision every hour.  
Overall this combinatorial explosion of commitment decisions has an adverse effect on ML models, since it creates distribution shifts on unseen commitments that is detrimental to the performance.
    
    \begin{figure}[!t]
        \centering
        \includegraphics[width=.5\columnwidth]{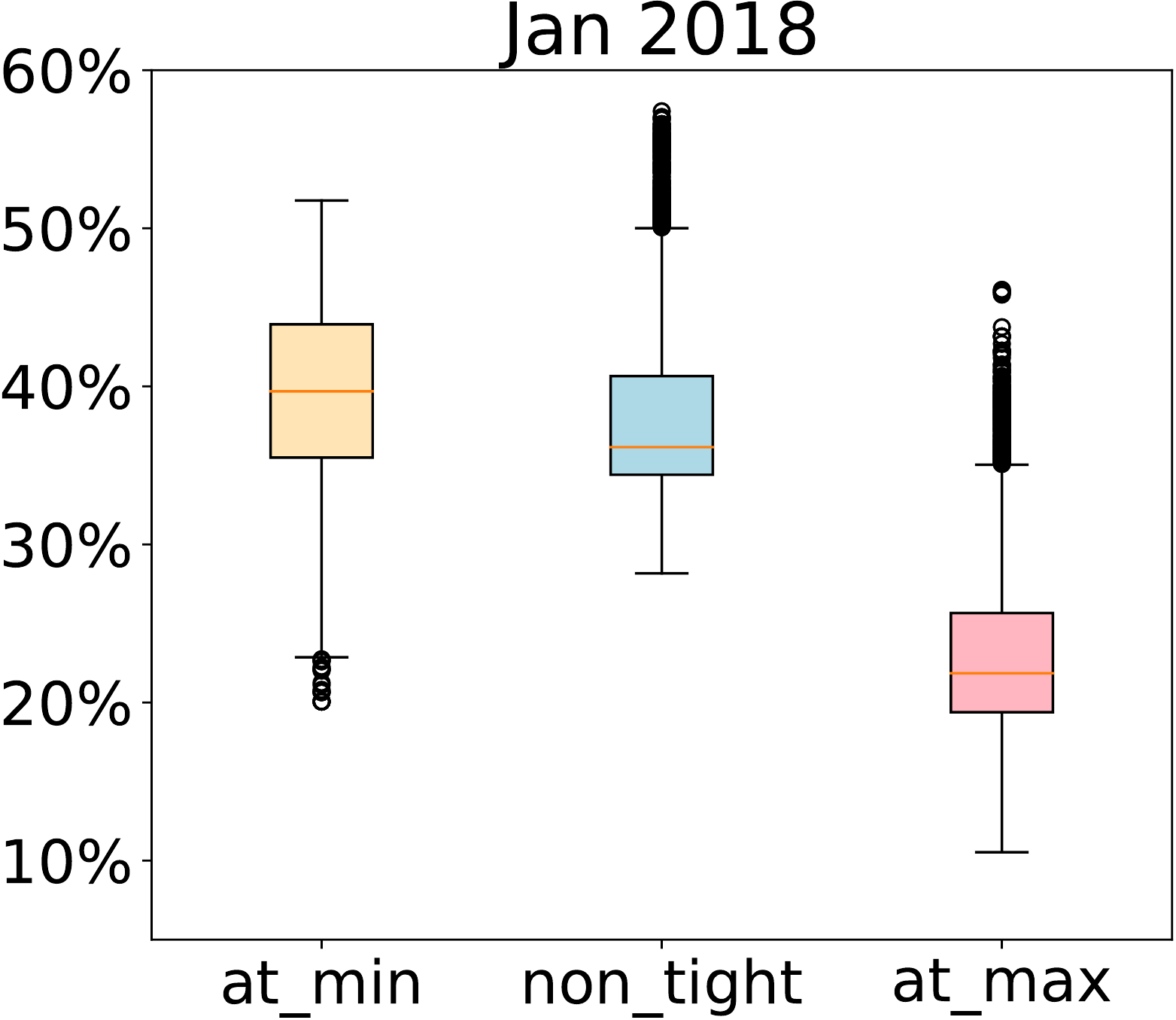}  
        \includegraphics[width=.438\columnwidth]{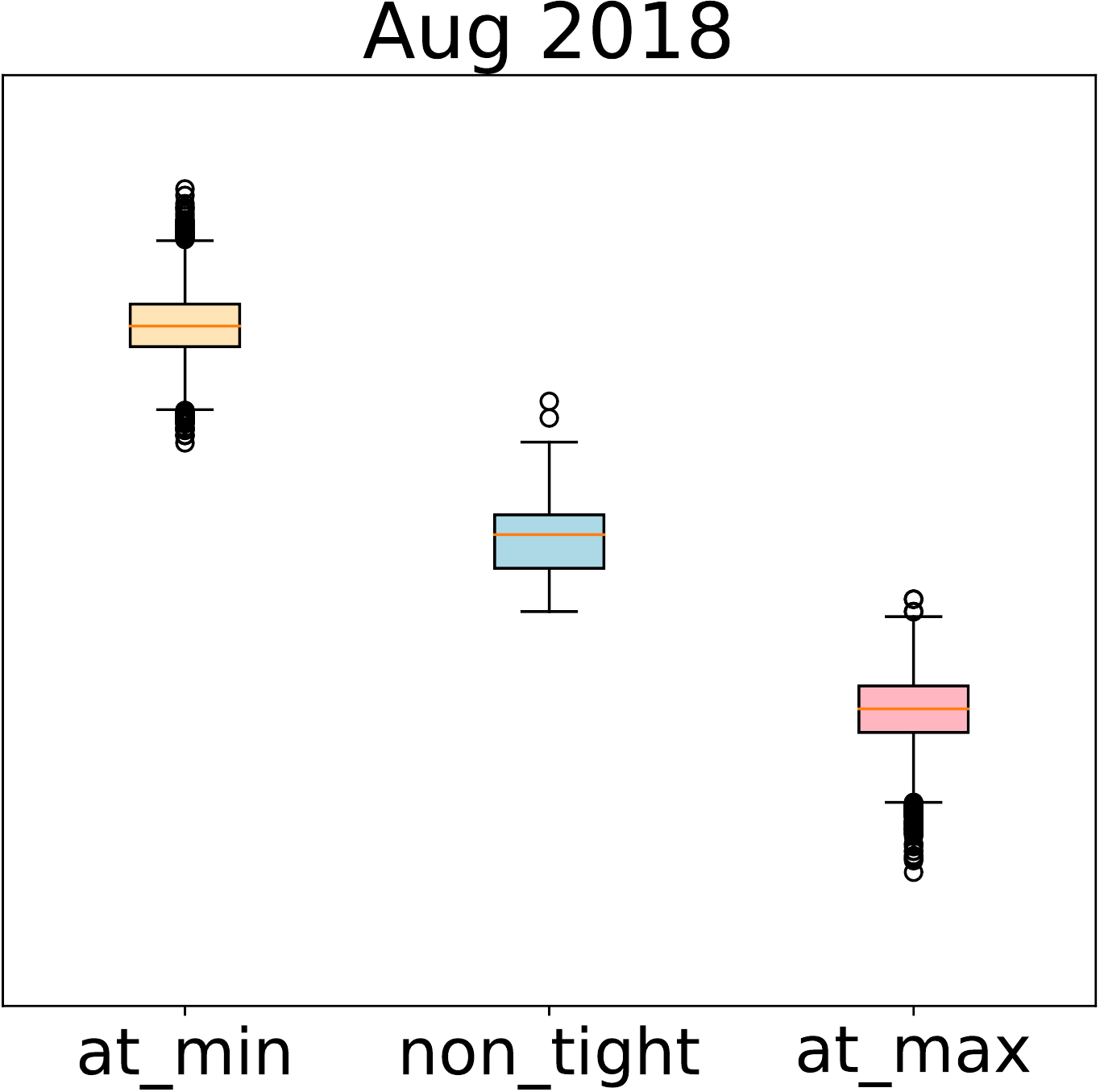}  
        \vspace{0.2cm}
        \caption{Proportion of generators at their maximum and minimum limits in January (left) and August (right).}
        \label{fig:dispatch_pattern}
    \end{figure}
    
Second, an analysis of RT-SCED solutions reveals, also unsurprisingly, that a majority of generators are dispatched to either their minimum or maximum limit; such limits include economic offer data and ramping constraints.  Detailed statistics are reported in Figure \ref{fig:dispatch_pattern} for January and August 2018: each plot quantifies, across all RT-SCED instances for that month, the proportion of generators dispatched at their minimum (at\_min) or maximum (at\_max) limit, or neither (non\_tight); these statistics exclude the renewable generators and the generators for which the minimum and maximum limit are identical.  In January, the median proportion of generators being dispatched at their minimum (resp., maximum) limit is close to $40\%$ (resp., $20\%$), while in August, these values are around $45\%$ and $20\%$.  
The variability is also visibly higher in winter, echoing the previous observations for
commitment decisions.

\subsection{First Classify, then Perform Regression}
\label{sec:ml_model}
    \begin{figure*}[!t]
        \centering
        \includegraphics[width=0.9\textwidth]{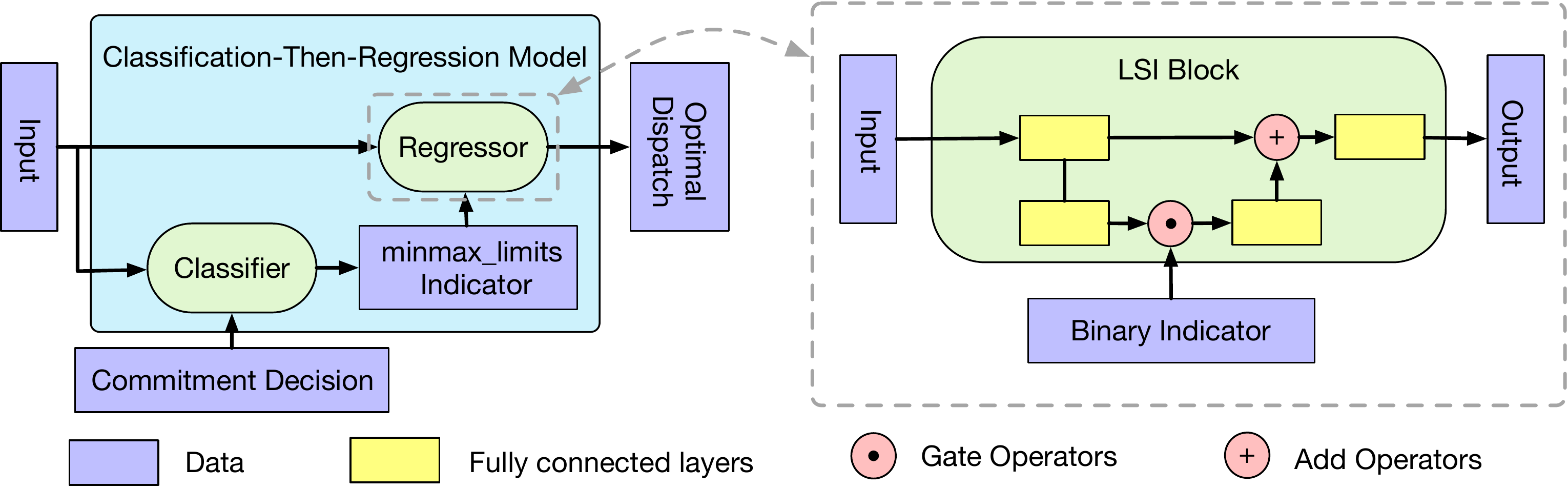}
        \vspace{0.2cm}
        \caption{The Proposed CTR model architecture. \textbf{Left}: Overall structure of the Classification-Then-Regression (CTR) model where the outputs of the classifier are used to inform the subsequent regressor. Thus the regressor only needs to predict the dispatches of non-tight generators; \textbf{Right}: The Latent Surgical Intervention (LSI) block for each classifier and regressor in the CTR model in which a gate operator takes the binary indicator to filter out some representation from fully connected layers. The result is then added into the representations element-wisely. }
        \label{fig:ML_model}
    \end{figure*}

    The previous analysis indicates that, given the knowledge of which generators are dispatched to their minimum or maximum limit, only a small number of generators need to be predicted.
    This suggests a \textbf{Classification-Then-Regression} (CTR) approach, first screening the active generators to identify those at their bounds, and then applying a regression to predict the dispatch of the remaining generators.
    The overall architecture is depicted in Figure~\ref{fig:ML_model}.
    The input of the learning task is a dataset $\mathcal{D} = \{(\mathbf{x}_i, \mathbf{y}_i, \mathbf{p}_i)\}_{i=1}^N$, where $\mathbf{x}_i$, $\mathbf{y}_i$, $\mathbf{p}_i$ represent the $i^{th}$ observation of the system state, the status indicators of the generators (i.e., whether each generator is at its maximum/minimum limits), and the optimal dispatches, respectively.

    \begin{table}[!t]
        \centering
        \caption{Input features of DNN model}
        \label{tab: ML_features}
        \begin{tabular}{ccc}
            \toprule
            Feature & Size & Source\\
            \midrule
            Loads & $L$ & Load forecasts\\
            Cost of generators & $G$ & Bids\\
            Cost of reserves & $2G$ & Bids\\
            Previous solution & $G$ & SCED\\
            Commitment decisions & $G$ & SCUC\\
            Reserve Commitments & $G$ & SCUC\\
            Generator min/max limits & $2G$ & Renewable forecasts\\
            Line losses factor & $2B+1$ & System\\
            \bottomrule
        \end{tabular}
    \end{table}

The CTR model is composed of two modules; classifier and
regressor. First, the classifier component aims at classifying whether
each generator is at its minimum or maximum limit. It considers the
status of the power network with $B$ buses, $G$ generators, and $L$
loads as its inputs.  Specifically, the classifier, parameterized by
$\mathbf{w}_1$, is a mapping $f_{\mathbf{w}_1}: \mathbb{R}^{d}
\xrightarrow{} \{0, 1\}^{2G}$, where $d$ is the dimension of the input
features. The overall input features describe the state of the power
system, detailed in Table~\ref{tab: ML_features}. Also, it outputs the
binary vector of $2G$ meaning that, for each generator, there are two
classification choices: one for determining whether it is dispatched at its
maximum limit and one for determining whether it is dispatched
at its minimum limit.  In the experiments presented in the subsequent
section, the dimension $d$ of the input space can be as large as
$24{,}403$ in the RTE system.  The optimal trainable parameters
$\mathbf{w^*_1}$ in the classifier are obtained through minimizing the
loss function as follows:
    
    \begin{align}
        \mathbf{w^*_1} = \argmin_{\mathbf{w_1}} \frac{1}{N}\sum_{i=1}^N \mathcal{L}_c(\mathbf{y}_i, f_{\mathbf{w}_1}(\mathbf{x}_i)), \label{eq: cls_loss}
    \end{align}
    where $\mathcal{L}_c$ denotes the cross entropy loss, i.e.,  
    \begin{align}
        \mathcal{L}_c(\mathbf{y}_i, \hat{\mathbf{y}}_i) = -\sum_{j=1}^{2G} \mathbf{y}_{i,j}\log(\hat{\mathbf{y}}_{i,j}) + (1-\mathbf{y}_{i,j})\log(1-\hat{\mathbf{y}}_{i,j}).
    \end{align}

    The second architectural component, the regressor that is parameterized by $\mathbf{w}_2$, is a mapping $f_{\mathbf{w}_2}: \mathbb{R}^{d+2G} \xrightarrow{} \mathbb{R}^{G}$.
    The additional $2G$ features in the input of the regressor come from the outputs of the classifier.
    Given the trained classifier $f_{\mathbf{w}^*_1}$, the optimal trainable parameters of the regressor $\mathbf{w^*_2}$ are obtained by minimizing the loss function $\mathcal{L}_r$ over all training instances as 
    \begin{align}
        \label{eq: reg_loss}
        \mathbf{w^*_2} = \argmin_{\mathbf{w_2}} \frac{1}{N}\sum_{i=1}^N \mathcal{L}_r \left( \mathbf{p}_i, f_{\mathbf{w}_2}\left(\mathbf{x}_i, f_{\mathbf{w}^*_1}(\mathbf{x}_i)\right)  \right), 
    \end{align}
    where $\mathcal{L}_r$ is the mean absolute error (MAE) loss, i.e., $\mathcal{L}_r(\mathbf{p}, \hat{\mathbf{p}}) = \|\mathbf{p} - \hat{\mathbf{p}}\|_1$.
    
The CTR architecture features a deep neural network
(DNN). Specifically, it uses a Latent Surgical Intervention (LSI)
network \cite{donnot2018latent} as its building block.  LSI is a
variant of residual neural networks \cite{he2016identity} that is
augmented by binary interventions.  As illustrated in Figure
\ref{fig:ML_model}, the LSI block exploits, via gate operators, the
binary information coming from commitment decisions and the
classifier.  As mentioned earlier, because the variability in SCED
solutions primarily comes from the combinatorial nature of the
commitment decisions, it is crucial to design the DNN architecture to
use commitment decisions as the input of the model.  Thus, the
LSI-based CTR model makes it possible to learn the generator dispatch
from various commitment decisions, thereby allowing the proposed
approach to generalize to the cases where the models are trained over
multiple commitment decisions.
    
\subsection{Machine Learning Pipeline} 
\label{sec:ML_pipeline}

The high variability of SCED instances in real operations, in terms of
commitment decisions and forecasts of loads and renewable energy
sources, makes it extremely challenging to learn the SCED optimization.
To mitigate this significant variability, this paper proposes a
learning pipeline that closely follows MISO's optimization pipeline.
The machine-learning pipeline is depicted in
Figure~\ref{fig:learning_pipeline} and consists of three phases: data
preparation, training, and prediction.
    
    \paragraph{Data Preparation} At 12pm of day $D-1$, the TSO outputs the commitment decisions of generators for the next 24 hours and Monte-Carlo scenarios for day $D$.
    Each scenario consists of 15 minute-level forecasts for load demands and renewable generators.
    Since the SCED is solved every 5 minutes throughout a day, the ML pipeline first linearly interpolates the forecasts at 5 minute level.
    Further data augmentation may be performed, if necessary, by perturbing load and renewable generations following the strategy described in \cite{chatzos2020high}.
    The data is then used as input to SCED optimization models, which generate the various optimal dispatches.
    The entire dataset is then divided into subsets, each of which spans one or a few hours, based on computational considerations. Indeed, the goal is to strike the proper balance between the accuracy of the models and the training time, since the models have to be available before midnight. This data preparation process yields the input instances to the learning task described previously.
    
\paragraph{Training}

First, each subset is split into the traditional
training/validation/test instances. The CTR models are trained on
training instances in sequence while the validation dataset is used
for hyperparameter tuning and the test dataset is used for reporting
the performance of the machine learning models. This training step
takes place in parallel for each subset.

\paragraph{Prediction} Starting from midnight on day $D$, at each time step, the corresponding ML model takes the latest system state as input and predicts the optimal dispatch of SCED models in real-time.

\section{Experimental Results}
\label{sec:results}

\subsection{Test Cases}
\label{sec:data_setup}

\begin{figure}
        \centering
        \includegraphics[width=0.48\columnwidth]{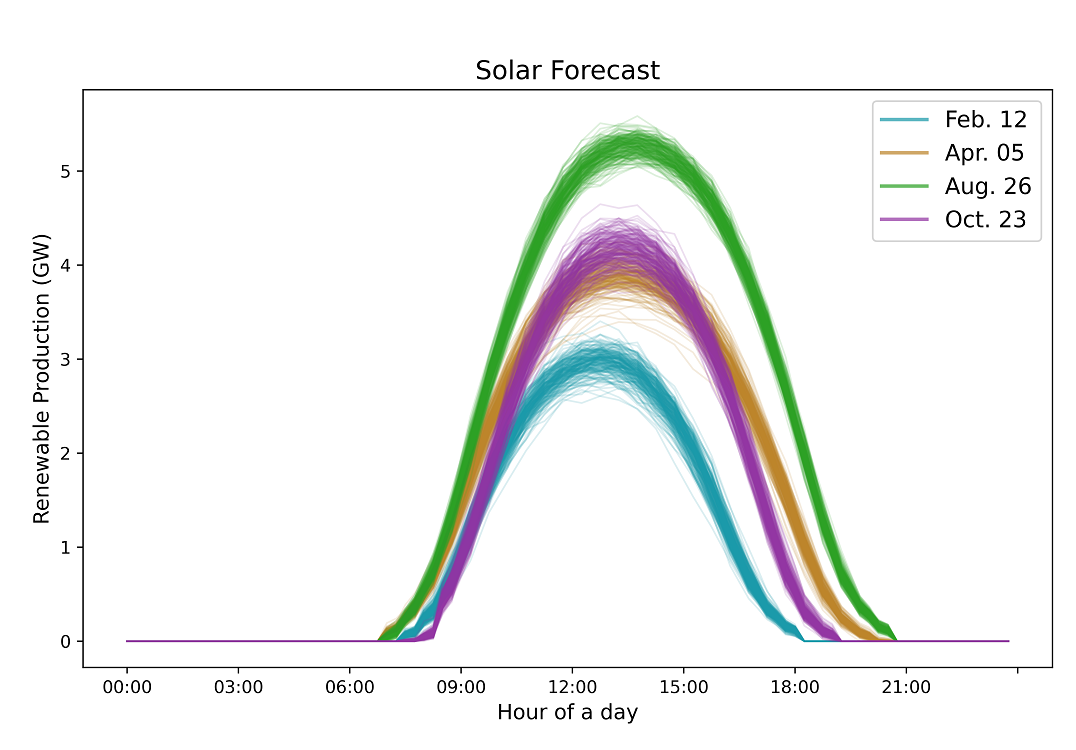}
        \includegraphics[width=0.48\columnwidth]{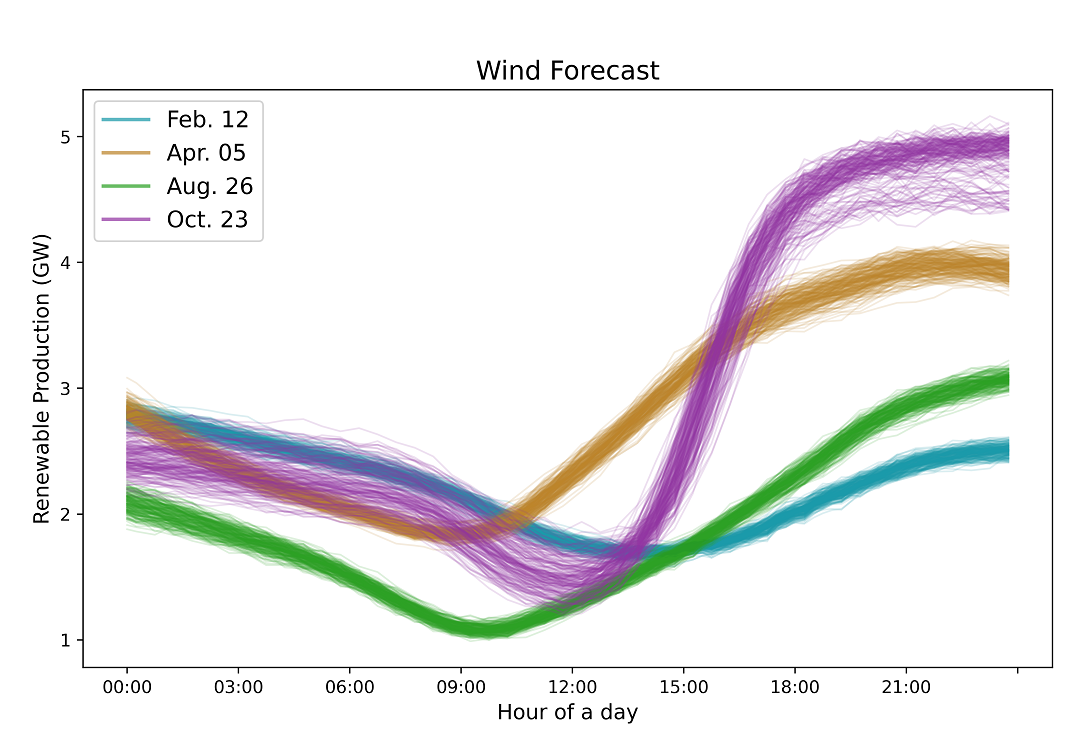}\\
        \includegraphics[width=0.48\columnwidth]{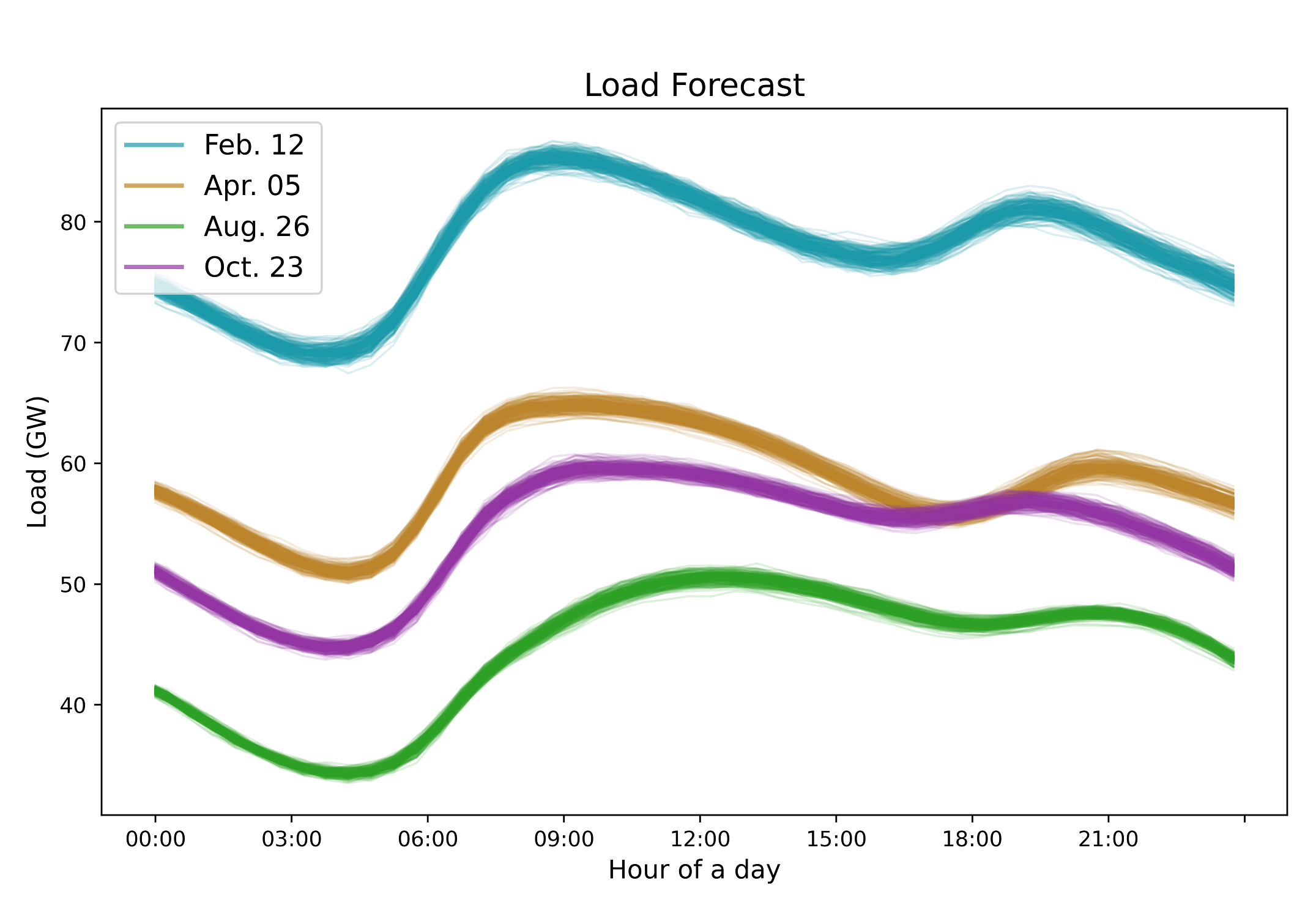}
        \includegraphics[width=0.48\columnwidth]{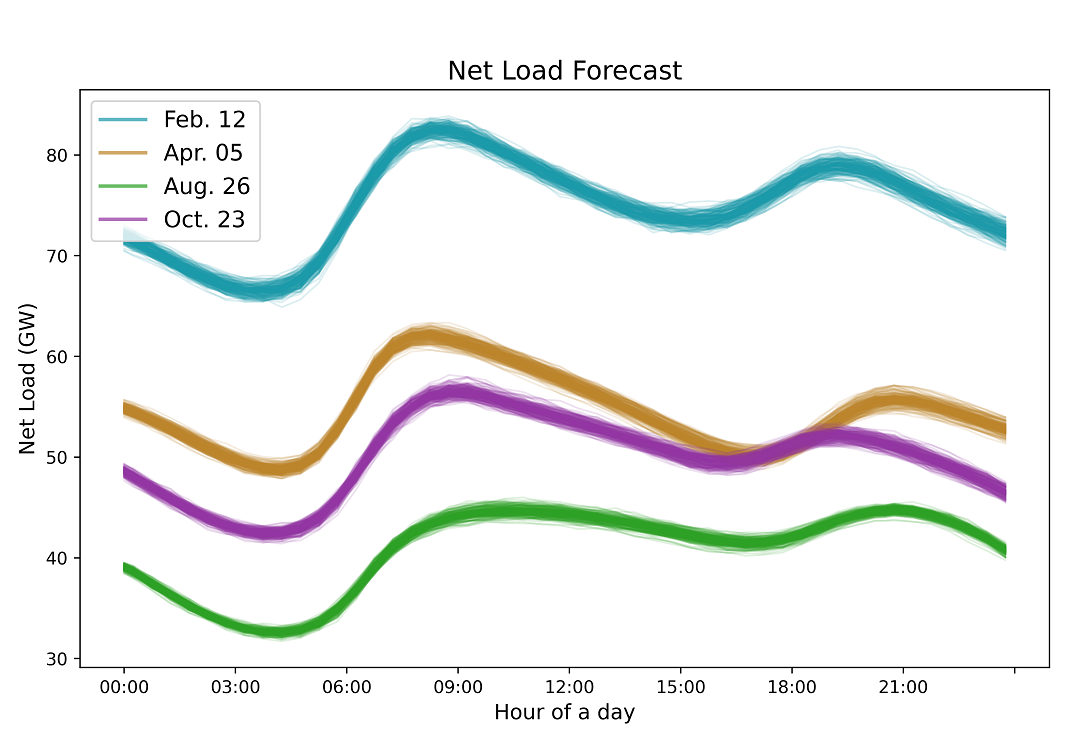}\\
        \caption{Day-ahead scenarios for solar output (top-left), wind output (top-right), total load (bottom-left) and net load (bottom-right). 200 Monte-Carlo scenarios are depicted.}
        \label{fig:res:scenarios}
    \end{figure}

The experiments replicate MISO's operations on the French transmission
system, for four representative days in 2018, namely, February
12\textsuperscript{th}, April 5\textsuperscript{th}, August
26\textsuperscript{th}, and October 23\textsuperscript{rd}.  These
four days were selected at random to represent annual seasonality.
For each operating day, 2000 Monte-Carlo scenarios for total load and
renewable power production are sampled in a day-ahead fashion, i.e.,
scenarios are produced at noon of the previous day.  These scenarios
are illustrated in Figure \ref{fig:res:scenarios}, which displays 200
scenarios for solar production, wind production, total load, and total
net load, respectively.  The total net load in this figure represents
the amount of power production expected to be generated by the
conventional generators, which is identical to the total load minus
the renewable generation.
    
The forecasting models for load consumption and renewable power
generation in this paper used Long Short-Term Memory (LSTM) neural
networks \cite{hochreiter1997long} and the scenarios are generated by
a MC-dropout approach \cite{gal2016dropout}. Other forecasting methods
and scenario generation approaches could be used instead, without
changing the methodology.

Day-ahead commitment decisions are obtained by solving a DA-SCUC problem and recording its solution.  
The SCUC formulation used in the present experiment follows
MISO's DA-SCUC described in \cite{BPM_002,BPM_002_B}.  
Again, the proposed ML pipeline is agnostic to the SCUC formulation itself and how it is solved, as it only requires the resulting commitments.

Finally, for each scenario and each day, 288 RT-SCED instances are
solved (one every 5 minutes), yielding a total of $576{,}000$
instances.  This initial dataset is then divided into 24 hourly
datasets, each containing $24{,}000$ SCED instances, since commitment
decisions are hourly.  To avoid information leakage, the data is split
between training/validation/testing instances as follows: $85\%$ of
scenarios are used for training, $7.5\%$ for validation, and $7.5\%$
for testing. All the reported results are in terms of the testing
instances.
    
The optimization problems (SCUC and SCED) are formulated in the JuMP
modeling language \cite{DunningHuchetteLubin2017_JuMP} and solved with
Gurobi 9.1 \cite{gurobi}, using Linux machines with dual Intel Xeon
6226@2.7GHz CPUs on the PACE Phoenix cluster \cite{PACE}; the entire
data generation phase is performed in less than 2 hours. The proposed
CTR model is implemented using PyTorch \cite{NEURIPS2019_9015} and trained using the Adam optimizer \cite{kingma2014adam} with a learning rate of 5e-4.
A grid search is performed to choose the hyperparameters, i.e., the
dimension of the hidden layers (taken in the set \{64, 128, 256\}),
and the number of layers (from the set \{2, 3, 4, 5, 6\}) for the
fully connected layers in the LSI block. A leaky ReLU with leakiness
of $\alpha=0.01$ is used as the activation function in the LSI
block. An early stopping criterion with 20 epochs is used for training the classifier and regressor models in order to prevent overfitting: when the loss values (Eq. \ref{eq: cls_loss} and Eq. \ref{eq: reg_loss}) on the validation dataset do not decrease for 20 consecutive epochs, the training process is terminated. Training is performed using Tesla V100-PCIE GPUs with 16GBs
HBM2 RAM, on machines with Intel CPU cores at 2.1GHz.

\subsection{Baselines}

The proposed CTR models are evaluated against the following baselines:
Naive CTR, Naive Regression (Naive Reg), and Regression (Reg). The
naive baselines replicate the behavior of the previous SCED solution,
i.e., they use the dispatch solution obtained 5 minutes earlier. The
naive baselines are motivated by the fact that, if the system does not
change much between two consecutive intervals, then the SCED solution
should not be changed much either.  The Naive CTR uses the same
approach as the CTR models: it first predicts whether a given
generator is at its minimum (resp., maximum) limit if it was at its
minimum (resp. maximum) limit in the previous dispatch; then, for the
remaining generators, it predicts the active dispatch using the
regressor. The naive baselines are expected to perform worse when
large fluctuations in load and renewable production are observed,
which typically occurs in the morning and evening.  Note that these
times of the day also display the largest ramping needs, and are among
the most critical for reliability. The effectiveness of the CTR
architecture is also demonstrated by comparing it to Reg for the optimal active dispatch, i.e., by omitting the
classification step from the CTR.

\subsection{Optimal Dispatch Prediction Errors}
        
    \begin{table}[!t]
        \centering
        \caption{Average Classification Accuracy (\%) of the CTR Classifier and Naive CTR on 4 Representative Days in 2018.}
        \label{tab:res:classification_overall}
        \begin{tabular}{lccccc}
            \toprule
            & \multicolumn{4}{c}{Dates} & \\
            \cmidrule{2-5}
            Methods & Feb. 12 & Apr. 5 & Aug. 26 & Oct. 23 & Avg. \\
            \midrule
            Naive classifier & 98.26 & 97.79 & 98.64 & 98.31 & 98.25 \\
            CTR & 99.56 & 99.18 & 99.40 & 99.24 & 99.35 \\
            \bottomrule
        \end{tabular}
    \end{table}

    Table \ref{tab:res:classification_overall} reports the overall
    classification accuracy of the CTR classifier and the Naive CTR across
    four representative days in 2018.  Surprisingly, the naive classifier
    is a strong baseline, with an accuracy that ranges from $97.79\%$ in
    the Spring to $98.64\%$ in the Summer.  The CTR classifier always
    improves on the baseline, by around one percentage point on average
    and by up to $1.40$ percentage point in the Spring. More detailed results
    about the classifiers are given in Appendix~\ref{appen:classification}.

\begin{figure}
    \centering
    \includegraphics[width=1\columnwidth]{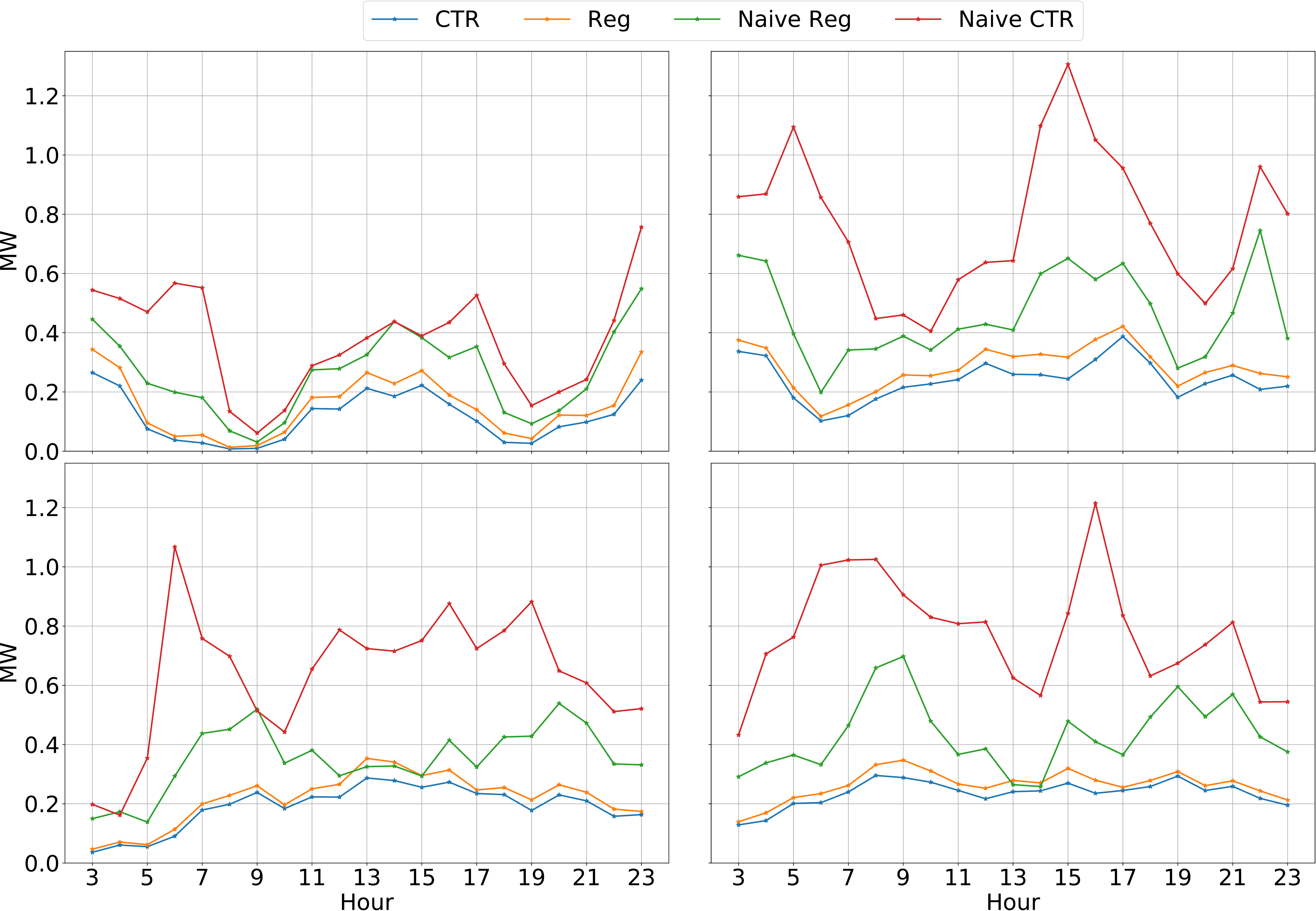}
    \caption{Mean Absolute Error (MAE) Over Time on Feb. 12 (top-left), Apr. 5 (top-right), Aug. 26 (bottom-left), and Oct. 23 (bottom-right).}
    \label{fig:MAE_along_time}
\end{figure}

\begin{table}[t]
    \centering
    \caption{Mean Absolute Error (MW) by Generator Size$^{\dagger}$.}
    \label{tab:res:disaggerated_dsipatch}
    \resizebox{0.95\columnwidth}{!}{
    \begin{tabular}{lrrrrrrrrrrrr}
        \toprule
         Date & Method & Small & Medium & Large & All \\
        \midrule
        \multirow{4}{*}{Feb. 12} & Naive Reg & 0.122 & 0.465 & 1.602 & 0.374 \\
                & Naive CTR & 0.084 & 0.333 & 1.128 & 0.262 \\
                & Reg & 0.057 & 0.188 & 0.654 & 0.153 \\
                & CTR & \textbf{0.043} & \textbf{0.141} & \textbf{0.535} & \textbf{0.117} \\
        \midrule
        \multirow{4}{*}{Apr. 05} & Naive Reg & 0.242 & 0.345 & 7.480 & 0.772 \\
                & Naive CTR & 0.197 & 0.220 & 4.374 & 0.463 \\
                & Reg & 0.149 & 0.152 & 2.553 & 0.282 \\
                & CTR & \textbf{0.105} & \textbf{0.110} & \textbf{2.291} & \textbf{0.241} \\
        \midrule
        \multirow{4}{*}{Aug. 26} & Naive Reg & 0.097 & 0.256 & 7.447 & 0.637 \\
                & Naive CTR & 0.080 & 0.149 & 4.045 & 0.352 \\
                & Reg & 0.054 & 0.124 & 2.454 & 0.218 \\
                & CTR & \textbf{0.034} & \textbf{0.064} & \textbf{2.220} & \textbf{0.190} \\
        \midrule
        \multirow{4}{*}{Oct. 23} & Naive Reg & 0.176 & 0.535 & 8.425 & 0.778 \\
                & Naive CTR & 0.140 & 0.341 & 4.596 & 0.434 \\
                & Reg & 0.106 & 0.192 & 2.724 & 0.263 \\
                & CTR & \textbf{0.076} & \textbf{0.145} & \textbf{2.525} & \textbf{0.235} \\
        \bottomrule
    \end{tabular}
    }
    \\
    $^{\dagger}$Small: 0-10MW; Medium: 10-100MW; Large: $>$100MW\\
\end{table}

Figure~\ref{fig:MAE_along_time} reports the mean absolute error (MAE)
for active power dispatch of the different models for each of the
considered days.  Given a ground-truth dispatch $p_i^g$
and the predicted dispatch $\hat{p}^g_i$, the MAE is
defined as
\begin{align*}
    MAE = \frac{1}{N} \frac{1}{G} \sum_{i=1}^{N} \sum_{g=1}^{G} \|p^g_i - \hat{p}^g_i\|,
\end{align*}
where $N$ is the number of instances in the test dataset and $G$ is
the number of generators for each instance.  As shown in
Figure~\ref{fig:MAE_along_time}, the MAEs of CTR are always lower than
those of Reg, demonstrating the benefits of the classifier. The MAEs
of Naive CTR are always lower than those of Naive Reg, showing that
even a naive classifier has benefits. Moreover, the methods using DNNs
for classification and regression, i.e., CTR and Reg, are always
better than their naive counterparts, demonstrating the value of deep
learning. Note that the performance of the naive methods fluctuates
during the day, mainly due to the variability of the commitments. The
naive methods are not robust with respect to changes in commitments,
contrary to the proposed CTR approach.

To investigate how the machine-learning models perform for different
generator types, Table~\ref{tab:res:disaggerated_dsipatch}
reports their behavior for different generator sizes.  The generators
are clustered into three groups based on their actual active
dispatches, and the MAE is reported for each group separately.
Small-size generators have a capacity between $0$ and $10$MW,
medium-size generators have a capacity between $10$ and $100$MW, and
large-size generators have a capacity above $100$MW.  The last column
report the MAEs across all
generators. Table~\ref{tab:res:disaggerated_dsipatch} further confirms
that the CTR models consistently outperform the corresponding
regressors.  These improvements are most notable for small and medium
generators which are expected to have higher variability: the MAEs are
decreased by up to $37.04\%$ across small generators and $48.39\%$
across medium generators (in Aug. 26).
Moreover, across four days, the CTR model achieves a Mean Average Percentage Error (MAPE) of $0.59\%$ and $0.34\%$ for medium and large generators.

\subsection{Solving Time vs Inference Time}
Solving the SCED using traditional optimization tools takes, on
average, $15.93$ seconds.  Actual computing times fluctuate throughout
the day, with increased load and congestion leading to higher
computing times.  In contrast, evaluating the optimization proxy for a
batch of 288 instances takes an average $1.5$ milliseconds.  In other
words, roughly $200{,}000$ scenarios may be evaluated in less than one
second.  This represents an improvement of 4 orders of magnitude, even
under the assumption that several hundred SCED instances can be solved
in parallel.

\section{Sensitivity Analysis}
\label{sec:results2}

\subsection{Motivations and Experiment Settings}

How many CTR models need to be trained is an important question to
ponder in practice. Preparing a single CTR model for 24 hours on Day
$D$ would be most convenient. However, as described in
Section~\ref{sec:pattern_dispatch}, the variability in commitments is
notoriously harmful to the prediction accuracy and training time of
the CTR model. However, successive hours during a day may only have a
few differences in commitments and hence a single CTR model may be
sufficient for predicting the associated SCED optimizations.
    
To answer that question, the original instance data is used to produce
a variety of datasets, containing data for consecutive 2, 3, 4, 6, 8,
12, and 24 hours.  For instance, the 8 hours dataset contains three
sets of instances, each grouping SCED data for 8 successive
hours. Three models are then trained using instances covering their 8
hours of data. All models use the termination criterion presented in Section \ref{sec:results}. 
The various CTR models for 1, 2, ..., 24 hours are
then compared for the 24 hours of the day, using the models
appropriate for each hour. It is expected that the quality of the
prediction will deteriorate with a coarser granularity, since there
will be a larger variability in commitments and net loads. However,
the LSI architecture of the CTR model and the availability of more
instances during training may compensate for this increase in
variability. In the following, CTR models trained on $i$ hours are
denoted as CTR$_i$.

\subsection{Results}

    \begin{figure*}[t]
    \centering
    \includegraphics[width=1.8\columnwidth]{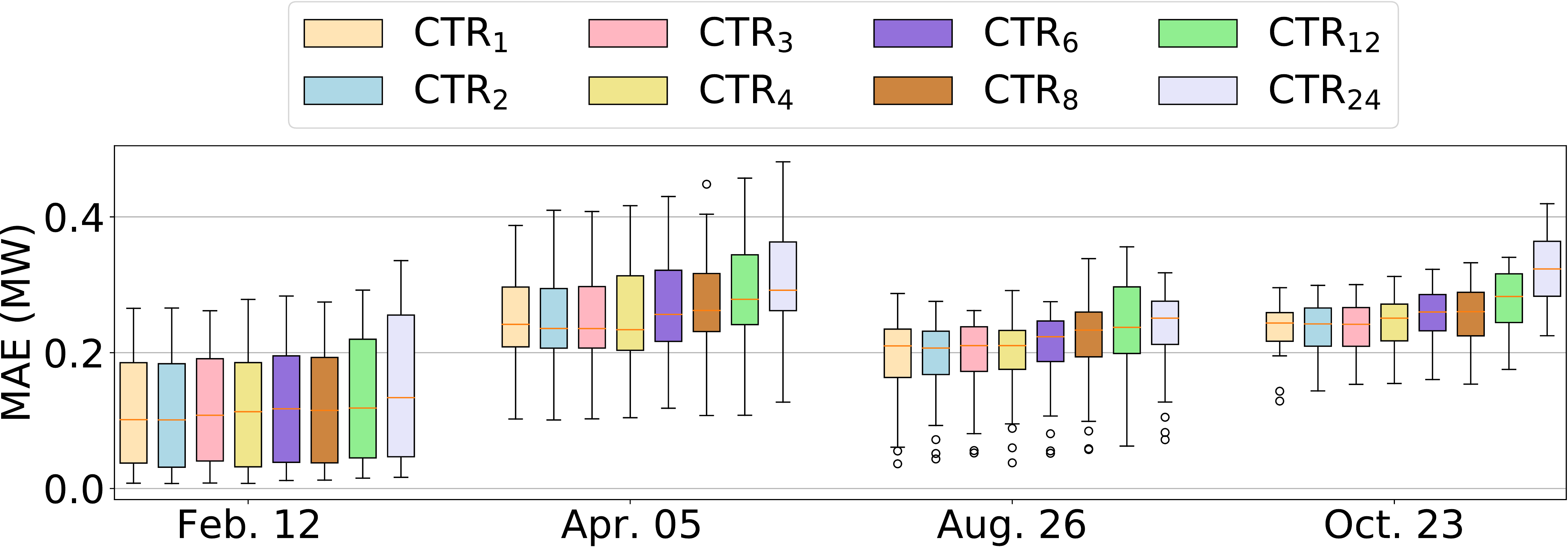}
    \caption{MAEs of CTR models for representative days: the CTR model
      scales well when trained over multiple hours. The performance
      degradation becomes significant only when aggregating 6 hours
      or more.}
    \label{fig:MAE_multiple_hours}
    \end{figure*}

Figure~\ref{fig:MAE_multiple_hours} illustrates the MAE values of the
various CTR models. The performance of the CTR model remains strong
even when aggregating up to 4 successive hours. As more hours are
aggregated, the performance starts to degrade. To get more insight on
the performance of the various CTR models, it is useful to consider
the energy distance \cite{rizzo2016energy} between the empirical
distributions of the testing and training instances. Recall that, given
two empirical distributions $\{x_i\}_{i=1}^N$ and $\{y_i\}_{i=1}^M$,
their energy distance is given by
    \begin{align*}
        \mathcal{E}(\mathcal{X},\mathcal{Y}) &= \frac{2}{NM} \sum_{i=1}^{N} \sum_{j=1}^{M} \|x_i - y_j\| \\
        &- \frac{1}{N^2} \sum_{i=1}^{N} \sum_{j=1}^{N} \|x_i - x_j\| \\
        &- \frac{1}{M^2} \sum_{i=1}^{M} \sum_{j=1}^{M} \|y_i - y_j\|.
    \end{align*}
    
    \begin{figure}[t]
    \centering
    \includegraphics[width=1\columnwidth]{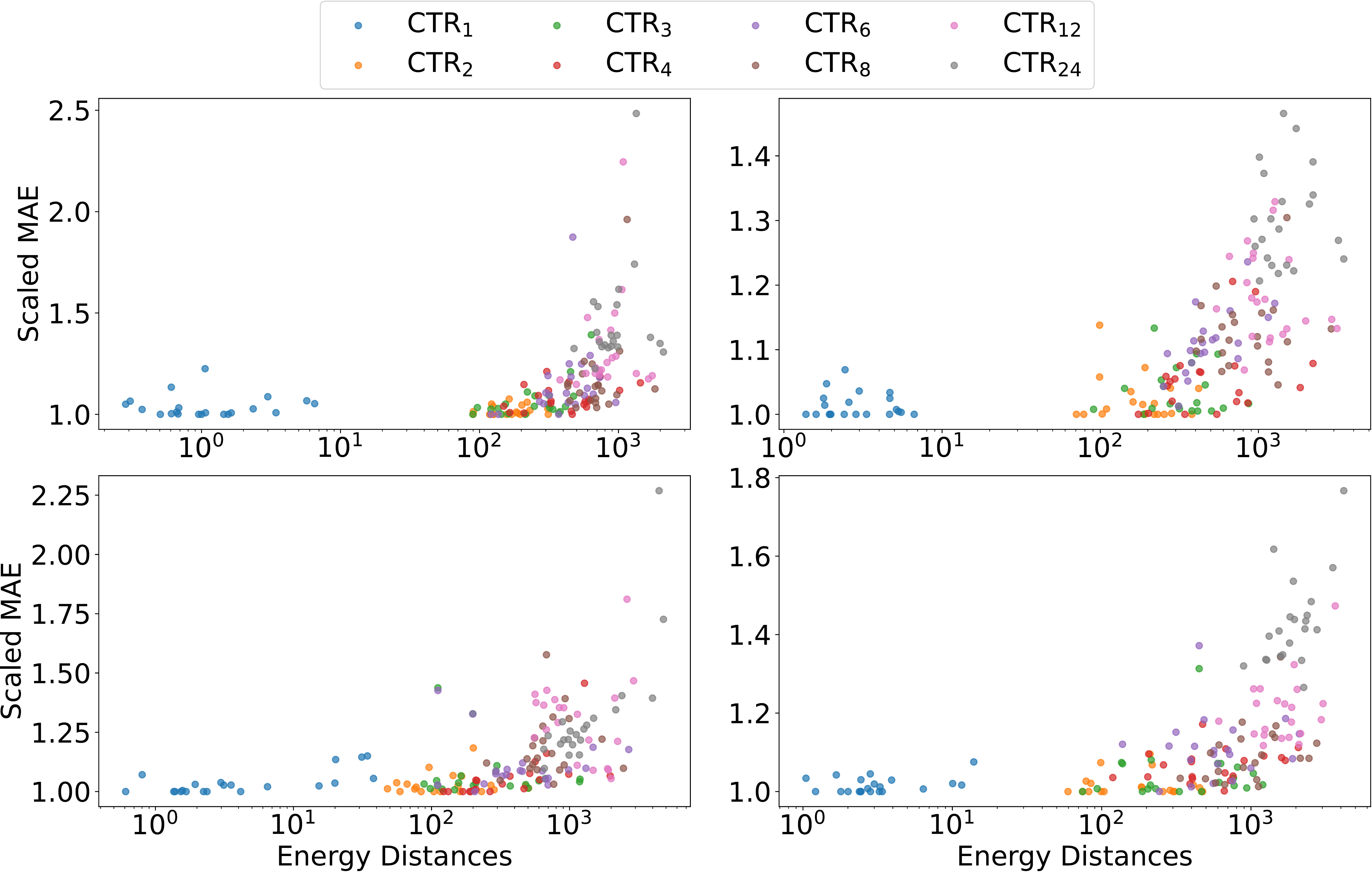}
    \caption{Scaled MAE value of the CTR models vs. energy distance between the test datasets and the training datasets for given hour's period on Feb. 12 (top-left), Apr. 5 (top-right), Aug. 26 (bottom-left), and Oct. 23 (bottom-right).}
    \label{fig:MAE_energy_distances}
    \end{figure}

It is thus interesting to study the relationship between the MAE of a
CTR model and the energy distance between its testing and training
datasets. Figure~\ref{fig:MAE_energy_distances} reports the
relationship between the scaled MAE and the energy distance, where the
scaled MAE is the MAE value divided by the best MAE value across all
experiments. Each dot captures the scaled MAE value (y-axis) for a CTR
model and the associated energy distance (x-axis) between its training
and testing datasets.

Observe first that the energy distance increases as more hourly data
are combined into the training set. In
Figure~\ref{fig:MAE_energy_distances}, the dots from CTR$_i$ tend to
have smaller energy distances than those of CTR$_j$ for $j > i$. The
scaled MAE also increases as the energy distance grows. More
interestingly, Figure~\ref{fig:MAE_energy_distances} shows that the
MAE value only increases slightly as the energy distance becomes
larger. Specifically, an increase of $10^2$ in energy distance
produces an increase of only 10\% in MAE (corresponding to a scaled MAE
of 1.1). These results confirm that the CTR models scale well when
facing a reasonable amount of variability in commitment decisions and
net loads.  However, when the energy distance goes over a certain
threshold, then the performance of CTR models starts to degrade
significantly even in presence of more data and more training time.
Note that the energy distance between the training and testing
datasets can be computed before the training process. Hence, the
proper granularity for the CTR model can be chosen to obtain the
desired accuracy. This quantification shows the potential of CTR to
perform well with more complex components involved in the MISO
pipeline such as the Look Ahead Commitment (LAC).
    
\begin{figure}[t]
    \centering
    \includegraphics[width=.49\columnwidth]{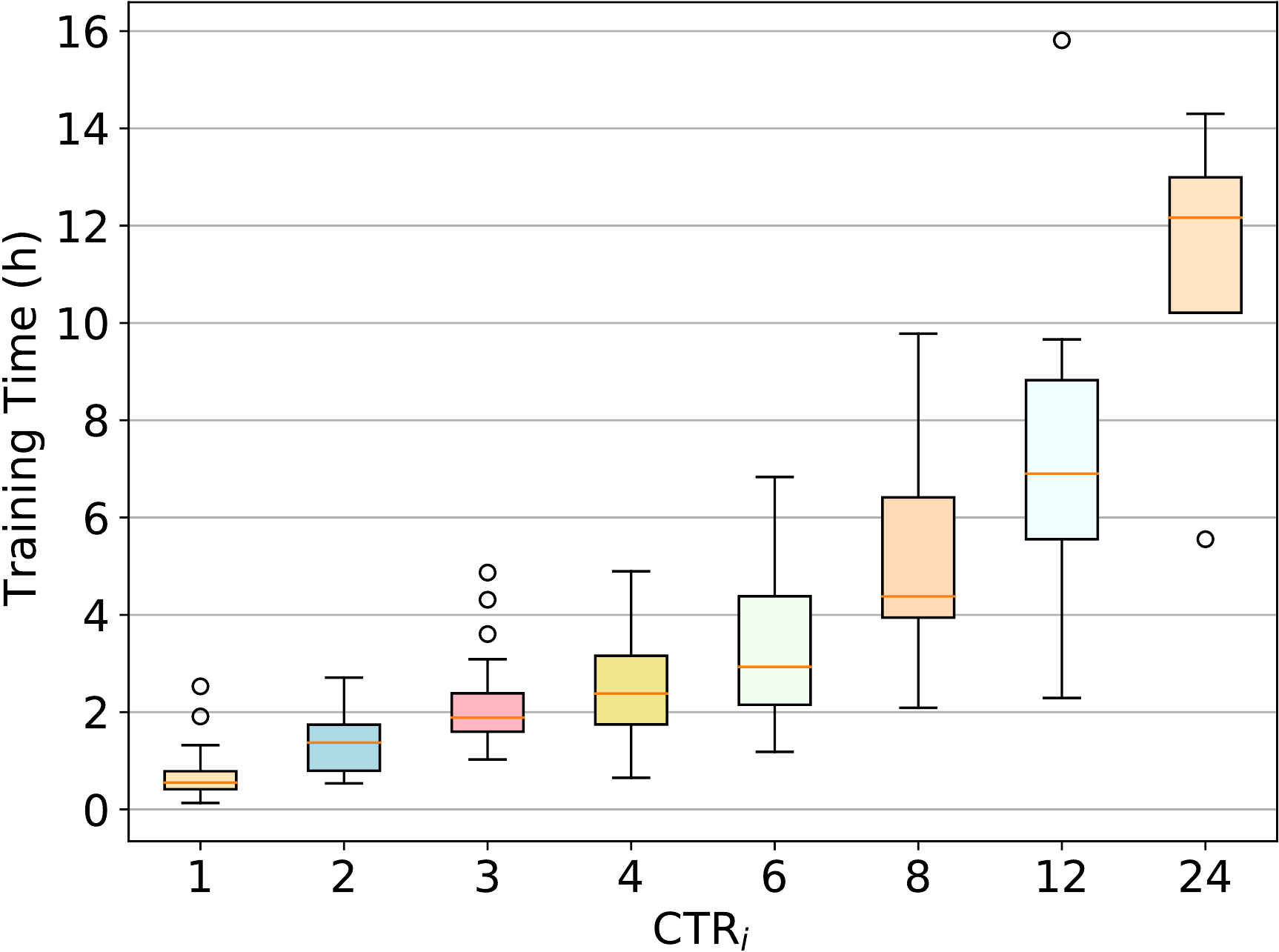}
    \includegraphics[width=.47\columnwidth]{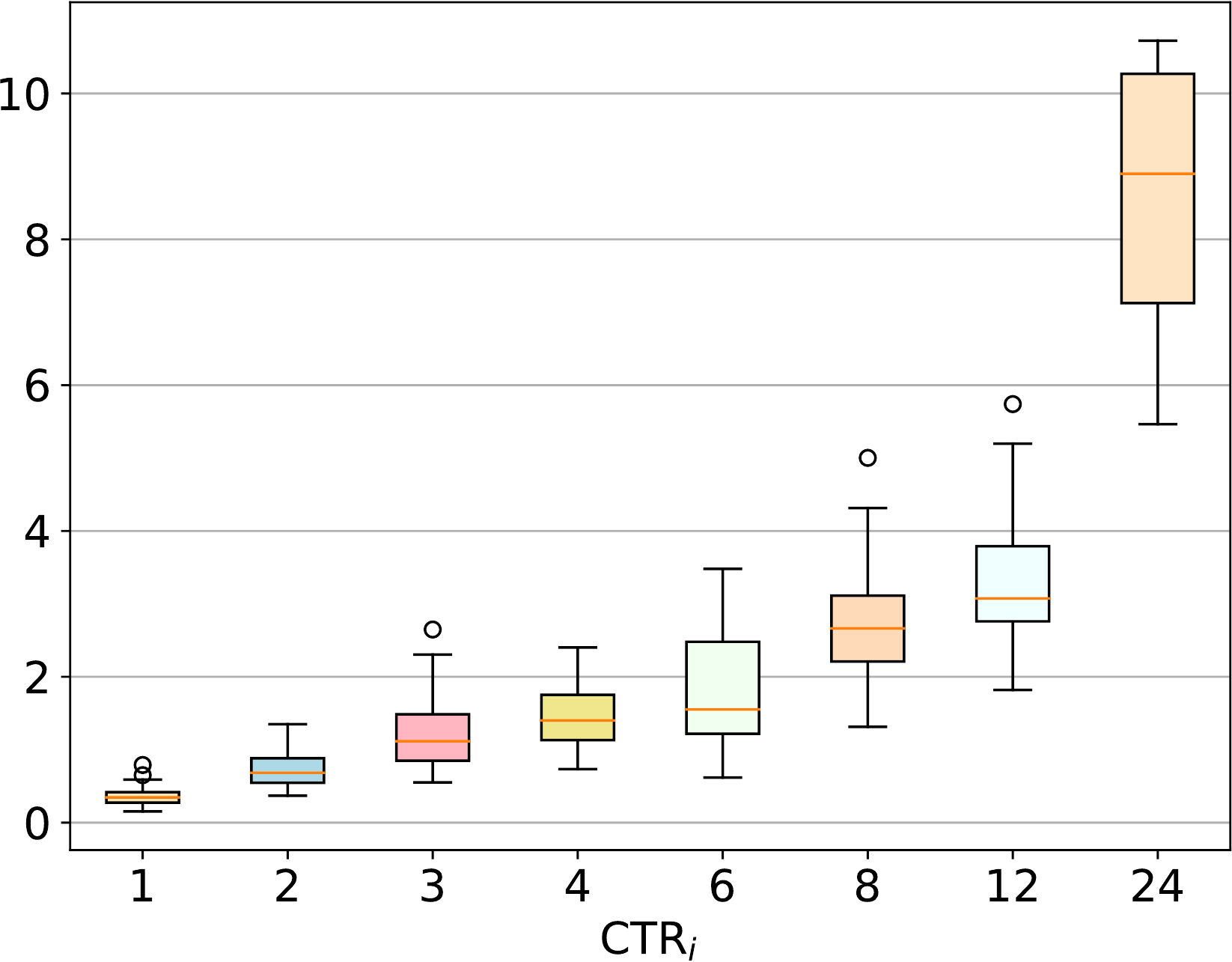}
    \caption{Training Times of CTR$_i$ Models: Regression (Left) and Classification (Right) }
    \label{fig:training_time}
    \end{figure}
    
Figure~\ref{fig:training_time} reports the training times of the CTR models. The average training time increases as more hourly data are included. The regression takes more training time than the classification task. Section \ref{sec:ML_pipeline} indicated that the training process must complete within about 12 hours after accumulating the data instances at 12pm on day $D-1$.
This is clearly possible for all the CTR$_i$ models with $i \leq 4$. Consider CTR$_4$ for instance. Figure~\ref{fig:training_time} shows that it can be trained for a 4-hour block within 10 hours in the worst case (4 hours for classification and 6 hours for regression). As a result, all 6 CRT$_4$ models can easily be trained in parallel within the required time frame. 

\section{Conclusion}

The paper presented several challenges that may hamper the use of ML
models in power systems operations, including the variability of
electricity demand and renewable production, the variations in
production costs, and the combinatorial structure of commitment
decisions.  To address these challenges, the paper proposed a novel ML
pipeline that leverages day-ahead forecasts and
the TSO's knowledge of commitment decisions several hours before they
take place.  The proposed pipeline consists of two main phases: 1)
day-ahead data preparation and training; and 2) real-time predictions,
which fits naturally within the operations of a TSO. Moreover, informed
by the behavior of real-time markets, the paper proposed a novel
Classification-Then-Regression (CTR) approach that leverages deep neural
networks based on Latent Surgical Interventions to capture generator
commitments and generators operating at their limits. Computational
experiments that replicated MISO's operational pipeline on a real,
large-scale transmission system demonstrated the feasibility of the
approach. In particular, the results show that optimization proxies
based on ML models have the potential to provide operators with new
tools for real-time risk monitoring.

Several extensions are possible, which will be the topic of future
work.  The integration, during training, of Lagrangian-based penalties
has the potential to further improve the performance of the neural
network models.  Such methods have been shown to improve the
feasibility of the predictions with respect to the original
constraints.  Moreover, once trained, the proposed classifier may be
used to also accelerate the SCED resolution, by providing hints as to
which variables should be fixed at their minimum or maximum limit.
Finally, the proposed CTR model is not limited to SCED models, and may
be applied to any optimization problems.

\section*{Acknowledgments}

This research is partly funded by NSF Awards 1912244
and 2112533, and ARPA-E Perform Award AR0001136.


\bibliographystyle{IEEEtran}
\bibliography{refs.bib}

\onecolumn
\appendix
\subsection{Classification accuracy across four days}\label{appen:classification}
    The detailed results of Naive and CTR model trained under different hourly data in four days are shown in Table~\ref{tab:res:classification:2018-02-12} \ref{tab:res:classification:2018-04-05} \ref{tab:res:classification:2018-08-26} \ref{tab:res:classification:2018-10-23}, respectively. First, across all days, CTR models always outperform the baseline model. Moreover, the performance decay is up to $3.74\%$ for Naive model (in Apr. 5) and up to $1.71\%$ for CTR model (CTR$_{12}$ in Aug. 26) along the day. CTR models have much smaller performance variations with the fluctuation of loads and renewables along the four days.
    Second, the CTR model scales well with more hours data included. The stark performance decay can only be observed with more than 12 hours data included.

    \begin{table*}[!hb]
        \centering
        \caption{Classification accuracy of Naive and CTR models (Feb. 12)}
        \resizebox{.7\columnwidth}{!}{%
        \label{tab:res:classification:2018-02-12}
        \begin{tabular}{lrrrrrrrrr}
            \toprule
            \multicolumn{1}{c}{Hour} & Naive & CTR$_1$ & CTR$_2$ & CTR$_3$ & CTR$_4$ & CTR$_6$ & CTR$_8$ & CTR$_{12}$ & CTR$_{24}$ \\
            \midrule
            03:00 - 04:00  & 98.37 & 99.24 & 99.24 & 99.29 & 99.24 & \textbf{99.31} & 99.26 & 99.17 & 98.94 \\
            04:00 - 05:00  & 98.79 & 99.36 & 99.38 & 99.40 & 99.32 & \textbf{99.42} & 99.37 & 99.30 & 99.11 \\
            05:00 - 06:00  & 98.74 & 99.68 & 99.70 & \textbf{99.71} & 99.69 & \textbf{99.71} & 99.70 & 99.65 & 99.55 \\
            06:00 - 07:00  & 98.60 & 99.85 & \textbf{99.87} & 99.86 & \textbf{99.87} & 99.86 & 99.86 & 99.84 & 99.77 \\
            07:00 - 08:00  & 98.14 & \textbf{99.88} & \textbf{99.88} & 99.87 & 99.85 & 99.87 & 99.85 & 99.85 & 99.80 \\
            08:00 - 09:00  & 98.66 & \textbf{99.97} & \textbf{99.97} & \textbf{99.97} & \textbf{99.97} & \textbf{99.97} & 99.96 & 99.96 & 99.95 \\
            09:00 - 10:00  & 99.27 & 99.96 & 99.96 & 99.96 & \textbf{99.97} & 99.96 & 99.96 & 99.95 & 99.94 \\
            10:00 - 11:00  & 99.04 & \textbf{99.80} & \textbf{99.80} & \textbf{99.80} & 99.79 & 99.77 & 99.77 & 99.74 & 99.70 \\
            11:00 - 12:00  & 97.64 & 99.24 & \textbf{99.38} & \textbf{99.38} & 99.34 & 99.29 & 99.31 & 99.21 & 99.08 \\
            12:00 - 13:00  & 97.96 & \textbf{99.37} & 99.36 & 99.34 & 99.31 & 99.30 & 99.31 & 99.19 & 99.04 \\
            13:00 - 14:00  & 97.97 & 99.22 & \textbf{99.26} & 99.25 & 99.21 & 99.21 & 99.18 & 99.05 & 98.91 \\
            14:00 - 15:00  & 97.25 & 99.29 & \textbf{99.32} & 99.30 & 99.29 & 99.29 & 99.27 & 99.13 & 98.95 \\
            15:00 - 16:00  & 98.37 & \textbf{99.25} & 99.23 & \textbf{99.25} & 99.18 & 99.19 & 99.16 & 99.06 & 98.94 \\
            16:00 - 17:00  & 98.71 & 99.52 & 99.53 & \textbf{99.54} & 99.53 & 99.51 & 99.45 & 99.43 & 99.34 \\
            17:00 - 18:00  & 98.12 & \textbf{99.69} & \textbf{99.69} & 99.68 & \textbf{99.69} & 99.65 & 99.62 & 99.59 & 99.53 \\
            18:00 - 19:00  & 98.93 & \textbf{99.88} & \textbf{99.88} & \textbf{99.88} & \textbf{99.88} & 99.82 & 99.83 & 99.82 & 99.78 \\
            19:00 - 20:00  & 99.17 & 99.87 & \textbf{99.88} & \textbf{99.88} & \textbf{99.88} & 99.84 & 99.85 & 99.83 & 99.81 \\
            20:00 - 21:00  & 99.03 & 99.68 & 99.64 & \textbf{99.69} & 99.59 & 99.58 & 99.59 & 99.58 & 99.54 \\
            21:00 - 22:00  & 97.83 & \textbf{99.47} & \textbf{99.47} & 99.39 & 99.38 & 99.31 & 99.38 & 99.33 & 99.22 \\
            22:00 - 23:00  & 96.11 & 99.45 & \textbf{99.47} & 99.42 & 99.37 & 99.32 & 99.36 & 99.30 & 99.23 \\
            23:00 - 24:00  & 96.82 & \textbf{99.02} & 98.96 & 98.89 & 98.82 & 98.76 & 98.81 & 98.71 & 98.60 \\
            \bottomrule
        \end{tabular}
        }
    \end{table*}
    \begin{table*}[!h]
        \centering
        \caption{Classification accuracy of Naive and CTR models (Apr. 05)}
        \resizebox{.7\columnwidth}{!}{%
        \label{tab:res:classification:2018-04-05}
        \begin{tabular}{lrrrrrrrrr}
            \toprule
            \multicolumn{1}{c}{Hour} & Naive & CTR$_1$ & CTR$_2$ & CTR$_3$ & CTR$_4$ & CTR$_6$ & CTR$_8$ & CTR$_{12}$ & CTR$_{24}$ \\
            \midrule
 03:00 - 04:00  & 97.95 & \textbf{99.24} & \textbf{99.24} & \textbf{99.24} & \textbf{99.24} & \textbf{99.24} & 99.21 & 99.11 & 98.95 \\
 04:00 - 05:00  & 98.29 & \textbf{99.34} & 99.33 & 99.33 & 99.31 & 99.33 & 99.31 & 99.23 & 99.08 \\
 05:00 - 06:00  & 98.43 & 99.44 & 99.45 & \textbf{99.46} & 99.44 & \textbf{99.46} & 99.44 & 99.36 & 99.19 \\
 06:00 - 07:00  & 99.02 & 99.66 & \textbf{99.69} & \textbf{99.69} & \textbf{99.69} & 99.60 & \textbf{99.69} & 99.65 & 99.55 \\
 07:00 - 08:00  & 98.20 & \textbf{99.56} & 99.54 & 99.55 & 99.53 & 99.46 & 99.53 & 99.49 & 99.35 \\
 08:00 - 09:00  & 98.22 & 99.31 & \textbf{99.34} & 99.30 & 99.32 & 99.24 & 99.23 & 99.24 & 99.11 \\
 09:00 - 10:00  & 98.05 & 99.27 & 99.27 & 99.27 & \textbf{99.28} & 99.15 & 99.16 & 99.17 & 98.97 \\
 10:00 - 11:00  & 98.40 & 99.26 & 99.29 & \textbf{99.30} & \textbf{99.30} & 99.21 & 99.23 & 99.23 & 99.08 \\
 11:00 - 12:00  & 97.68 & 98.94 & 98.97 & 98.98 & \textbf{99.00} & 98.86 & 98.89 & 98.88 & 98.64 \\
 12:00 - 13:00  & 97.11 & 98.46 & \textbf{98.51} & 98.45 & 98.44 & 98.22 & 98.42 & 98.10 & 98.05 \\
 13:00 - 14:00  & 97.36 & \textbf{98.83} & 98.80 & 98.77 & 98.78 & 98.56 & 98.65 & 98.45 & 98.32 \\
 14:00 - 15:00  & 96.64 & 98.94 & 98.97 & 98.98 & \textbf{99.00} & 98.82 & 98.89 & 98.71 & 98.64 \\
 15:00 - 16:00  & 96.55 & 99.10 & \textbf{99.21} & 99.14 & \textbf{99.21} & 99.01 & 99.07 & 98.93 & 98.81 \\
 16:00 - 17:00  & 97.36 & \textbf{98.93} & 98.71 & 98.81 & 98.56 & 98.60 & 98.54 & 98.52 & 98.38 \\
 17:00 - 18:00  & 97.03 & \textbf{98.72} & 98.59 & 98.67 & 98.51 & 98.46 & 98.47 & 98.42 & 98.31 \\
 18:00 - 19:00  & 98.01 & 99.15 & \textbf{99.19} & \textbf{99.19} & 99.02 & 99.11 & 99.02 & 98.95 & 98.88 \\
 19:00 - 20:00  & 98.72 & 99.33 & 99.39 & \textbf{99.40} & 99.21 & 99.33 & 99.27 & 99.20 & 99.15 \\
 20:00 - 21:00  & 98.87 & 99.37 & 99.41 & \textbf{99.42} & 99.38 & 99.38 & 99.31 & 99.24 & 99.21 \\
 21:00 - 22:00  & 98.04 & 99.41 & \textbf{99.43} & 99.34 & 99.39 & 99.35 & 99.30 & 99.26 & 99.22 \\
 22:00 - 23:00  & 95.28 & 99.22 & 99.23 & 99.23 & \textbf{99.25} & 99.17 & 99.07 & 99.00 & 98.87 \\
 23:00 - 24:00  & 98.37 & \textbf{99.39} & 99.36 & 99.35 & 99.37 & 99.33 & 99.25 & 99.22 & 99.16 \\
            \bottomrule
        \end{tabular}
        }
    \end{table*}
    
    \vspace{10cm}
    
    \begin{table*}[!h]
        \centering
        \caption{Classification accuracy of Naive and CTR models (Aug. 26)}
        \resizebox{.7\columnwidth}{!}{%
        \label{tab:res:classification:2018-08-26}
        \begin{tabular}{lrrrrrrrrr}
            \toprule
            \multicolumn{1}{c}{Hour} & Naive & CTR$_1$ & CTR$_2$ & CTR$_3$ & CTR$_4$ & CTR$_6$ & CTR$_8$ & CTR$_{12}$ & CTR$_{24}$ \\
            \midrule
 03:00 - 04:00  & 99.73 & \textbf{99.95} & 99.94 & 99.92 & \textbf{99.95} & 99.92 & 99.91 & 99.90 & 99.88 \\
 04:00 - 05:00  & 99.73 & \textbf{99.92} & 99.91 & 99.89 & 99.88 & 99.89 & 99.88 & 99.88 & 99.85 \\
 05:00 - 06:00  & 99.66 & 99.88 & \textbf{99.90} & 99.88 & 99.87 & 99.88 & 99.87 & 99.86 & 99.84 \\
 06:00 - 07:00  & 97.93 & 99.65 & \textbf{99.67} & 99.66 & \textbf{99.67} & 99.62 & 99.65 & 99.61 & 99.50 \\
 07:00 - 08:00  & 98.94 & \textbf{99.63} & \textbf{99.63} & \textbf{99.63} & \textbf{99.63} & 99.60 & 99.60 & 99.59 & 99.51 \\
 08:00 - 09:00  & 98.94 & \textbf{99.62} & \textbf{99.62} & \textbf{99.62} & 99.59 & 99.58 & 99.46 & 99.58 & 99.49 \\
 09:00 - 10:00  & 98.87 & 99.52 & \textbf{99.55} & 99.49 & 99.52 & 99.51 & 99.39 & 99.52 & 99.41 \\
 10:00 - 11:00  & 98.91 & \textbf{99.59} & 99.58 & 99.57 & 99.58 & 99.57 & 99.50 & 99.58 & 99.50 \\
 11:00 - 12:00  & 97.98 & 99.15 & 99.18 & \textbf{99.19} & \textbf{99.19} & 99.17 & 99.15 & 99.15 & 99.10 \\
 12:00 - 13:00  & 98.08 & 99.03 & \textbf{99.07} & 99.06 & \textbf{99.07} & 98.95 & 98.83 & 98.44 & 98.82 \\
 13:00 - 14:00  & 98.31 & 98.71 & 98.87 & 98.97 & \textbf{98.98} & 98.81 & 98.64 & 98.35 & 98.65 \\
 14:00 - 15:00  & 98.05 & 98.68 & 98.84 & 98.93 & \textbf{98.97} & 98.90 & 98.56 & 98.20 & 98.60 \\
 15:00 - 16:00  & 98.15 & 98.77 & 98.90 & 98.89 & \textbf{98.98} & 98.90 & 98.67 & 98.29 & 98.68 \\
 16:00 - 17:00  & 97.85 & 98.75 & 98.75 & 98.88 & 98.62 & \textbf{98.94} & 98.32 & 98.19 & 98.66 \\
 17:00 - 18:00  & 98.96 & 99.32 & 99.36 & \textbf{99.38} & \textbf{99.38} & 99.31 & 99.37 & 99.11 & 99.31 \\
 18:00 - 19:00  & 98.77 & 99.42 & \textbf{99.46} & 99.44 & 99.43 & 99.36 & 99.39 & 99.10 & 99.28 \\
 19:00 - 20:00  & 98.68 & \textbf{99.64} & \textbf{99.64} & 99.63 & 99.61 & 99.59 & 99.58 & 99.42 & 99.53 \\
 20:00 - 21:00  & 98.42 & \textbf{99.55} & 99.50 & 99.50 & 99.53 & 99.46 & 99.46 & 99.25 & 99.38 \\
 21:00 - 22:00  & 97.96 & 99.46 & 99.44 & 99.45 & \textbf{99.47} & 99.35 & 99.36 & 99.06 & 99.25 \\
 22:00 - 23:00  & 98.63 & 99.59 & 99.61 & 99.59 & \textbf{99.62} & 99.52 & 99.53 & 99.31 & 99.46 \\
 23:00 - 24:00  & 98.89 & 99.63 & \textbf{99.64} & 99.61 & 99.62 & 99.52 & 99.54 & 99.29 & 99.46 \\
            \bottomrule
        \end{tabular}
        }
    \end{table*}
    \begin{table*}[!h]
        \centering
        \caption{Classification accuracy of Naive and CTR models (Oct. 23)}
        \resizebox{.7\columnwidth}{!}{%
        \label{tab:res:classification:2018-10-23}
        \begin{tabular}{lrrrrrrrrr}
            \toprule
            \multicolumn{1}{c}{Hour} & Naive & CTR$_1$ & CTR$_2$ & CTR$_3$ & CTR$_4$ & CTR$_6$ & CTR$_8$ & CTR$_{12}$ & CTR$_{24}$ \\
            \midrule
 03:00 - 04:00  & 99.46 & \textbf{99.75} & \textbf{99.75} & 99.71 & \textbf{99.75} & 99.70 & 99.71 & 99.64 & 99.60 \\
 04:00 - 05:00  & 98.62 & 99.55 & \textbf{99.56} & 99.50 & 99.54 & 99.50 & 99.54 & 99.39 & 99.22 \\
 05:00 - 06:00  & 98.95 & 99.57 & \textbf{99.60} & 99.55 & 99.58 & 99.55 & \textbf{99.60} & 99.50 & 99.32 \\
 06:00 - 07:00  & 99.16 & 99.55 & \textbf{99.56} & 99.53 & 99.55 & 99.42 & 99.55 & 99.42 & 99.29 \\
 07:00 - 08:00  & 98.43 & 99.31 & \textbf{99.34} & 99.30 & 99.25 & 99.09 & 99.27 & 99.01 & 98.74 \\
 08:00 - 09:00  & 97.02 & 98.99 & 98.99 & \textbf{99.01} & 98.90 & 98.80 & 98.70 & 98.67 & 98.41 \\
 09:00 - 10:00  & 97.14 & 99.03 & \textbf{99.05} & 98.90 & 98.91 & 98.83 & 98.71 & 98.65 & 98.33 \\
 10:00 - 11:00  & 97.66 & \textbf{98.85} & 98.67 & 98.75 & 98.77 & 98.67 & 98.55 & 98.50 & 98.15 \\
 11:00 - 12:00  & 97.74 & \textbf{98.75} & 98.62 & 98.71 & 98.68 & 98.65 & 98.60 & 98.48 & 98.24 \\
 12:00 - 13:00  & 97.34 & 98.61 & \textbf{98.70} & 98.65 & 98.52 & 98.41 & 98.51 & 98.26 & 98.05 \\
 13:00 - 14:00  & 98.43 & 98.81 & 98.84 & \textbf{98.86} & 98.71 & 98.68 & 98.66 & 98.55 & 98.40 \\
 14:00 - 15:00  & 98.33 & 98.76 & \textbf{98.86} & 98.82 & 98.70 & 98.62 & 98.63 & 98.52 & 98.35 \\
 15:00 - 16:00  & 97.00 & 98.64 & \textbf{98.83} & 98.76 & 98.63 & 98.57 & 98.52 & 98.33 & 98.07 \\
 16:00 - 17:00  & 98.47 & 99.31 & 99.33 & \textbf{99.36} & 99.31 & 99.15 & 99.28 & 99.08 & 98.99 \\
 17:00 - 18:00  & 99.13 & \textbf{99.47} & \textbf{99.47} & \textbf{99.47} & \textbf{99.47} & 99.33 & 99.46 & 99.32 & 99.25 \\
 18:00 - 19:00  & 99.13 & 99.57 & \textbf{99.58} & 99.57 & 99.54 & 99.55 & 99.54 & 99.43 & 99.36 \\
 19:00 - 20:00  & 98.20 & 99.35 & \textbf{99.42} & \textbf{99.42} & 99.34 & 99.38 & 99.36 & 99.21 & 99.06 \\
 20:00 - 21:00  & 98.78 & 99.48 & 99.51 & \textbf{99.52} & 99.47 & 99.49 & 99.47 & 99.34 & 99.20 \\
 21:00 - 22:00  & 97.76 & \textbf{99.54} & 99.53 & 99.52 & 99.52 & 99.52 & 99.50 & 99.40 & 99.18 \\
 22:00 - 23:00  & 98.69 & \textbf{99.59} & \textbf{99.59} & 99.58 & 99.58 & 99.57 & 99.56 & 99.46 & 99.40 \\
 23:00 - 24:00  & 99.01 & 99.57 & \textbf{99.58} & 99.55 & 99.55 & 99.53 & 99.53 & 99.43 & 99.34 \\
            \bottomrule
        \end{tabular}
        }
    \end{table*}

\newpage
\subsection{Hourly data distribution}\label{appen:data_dist}
\begin{figure*}[!h]
\centering
\includegraphics[width=0.30\linewidth]{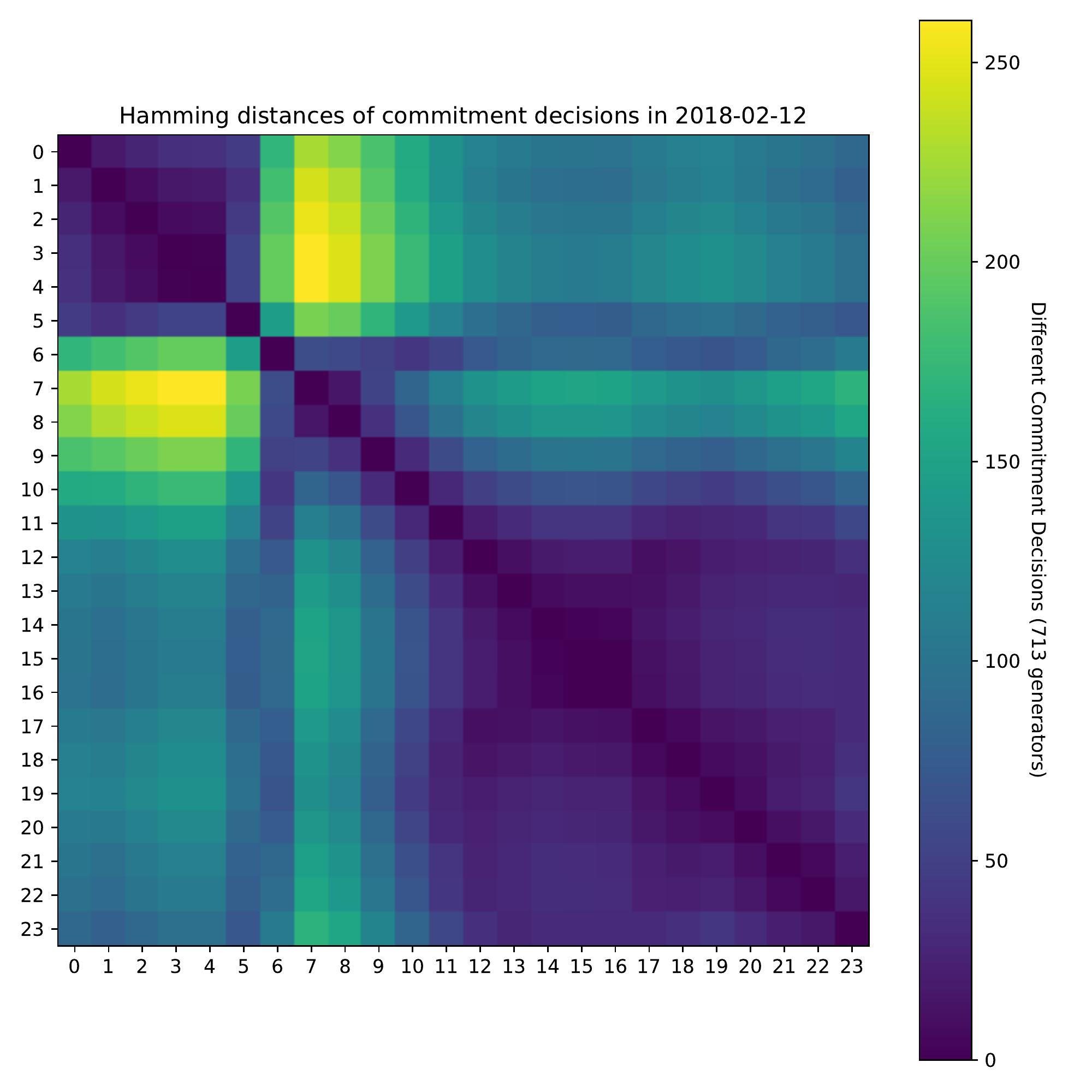}
\includegraphics[width=0.30\linewidth]{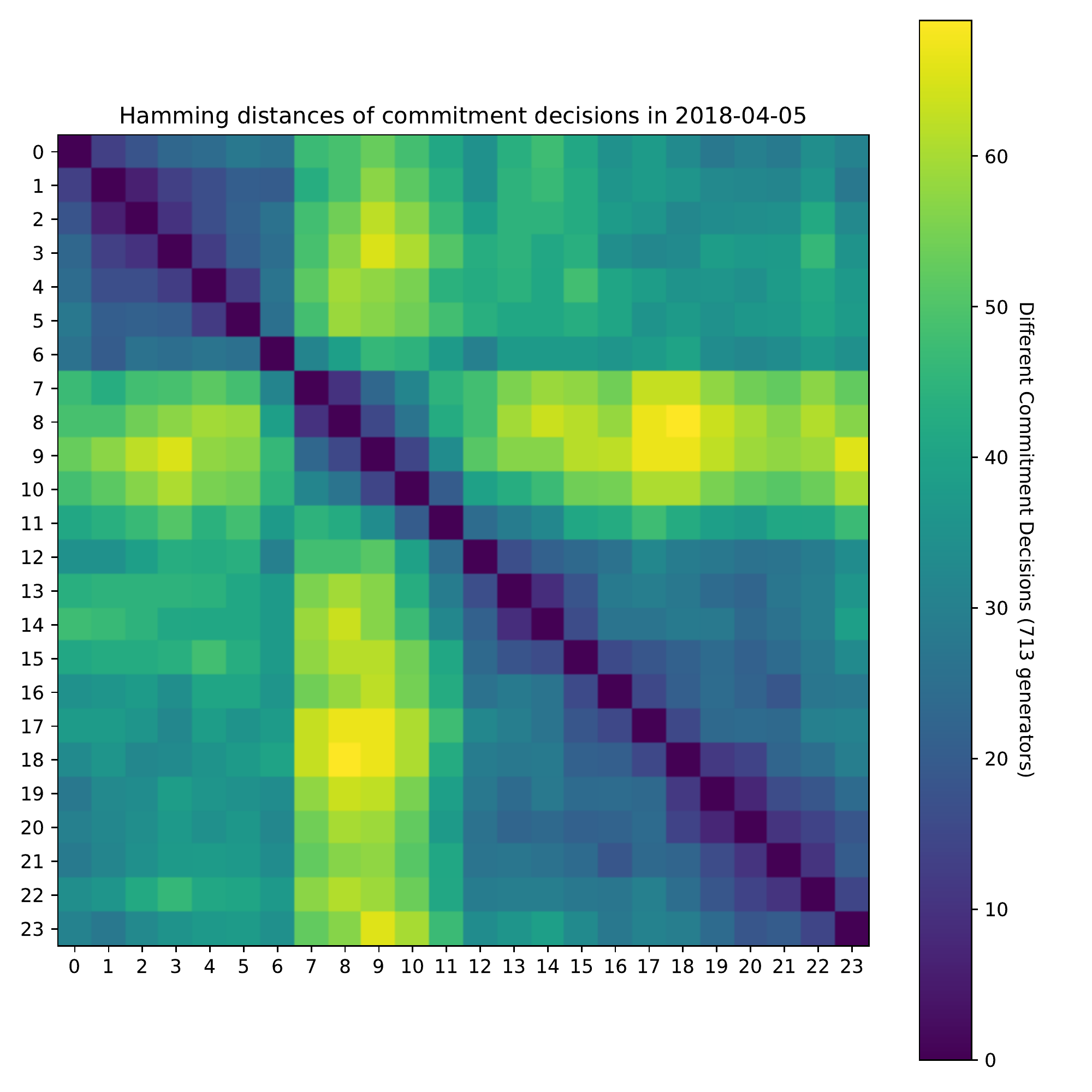}\\
\includegraphics[width=0.30\linewidth]{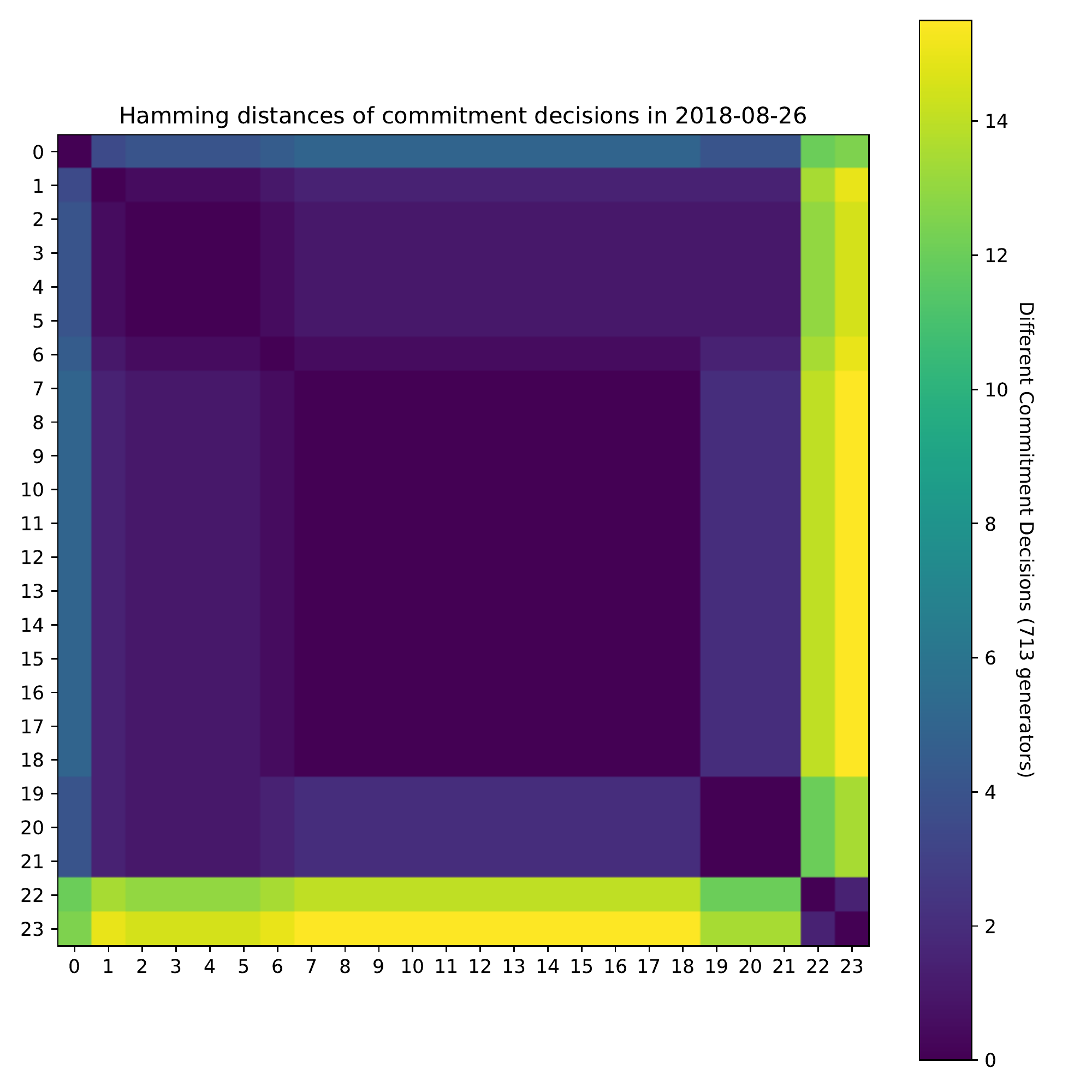}
\includegraphics[width=0.30\linewidth]{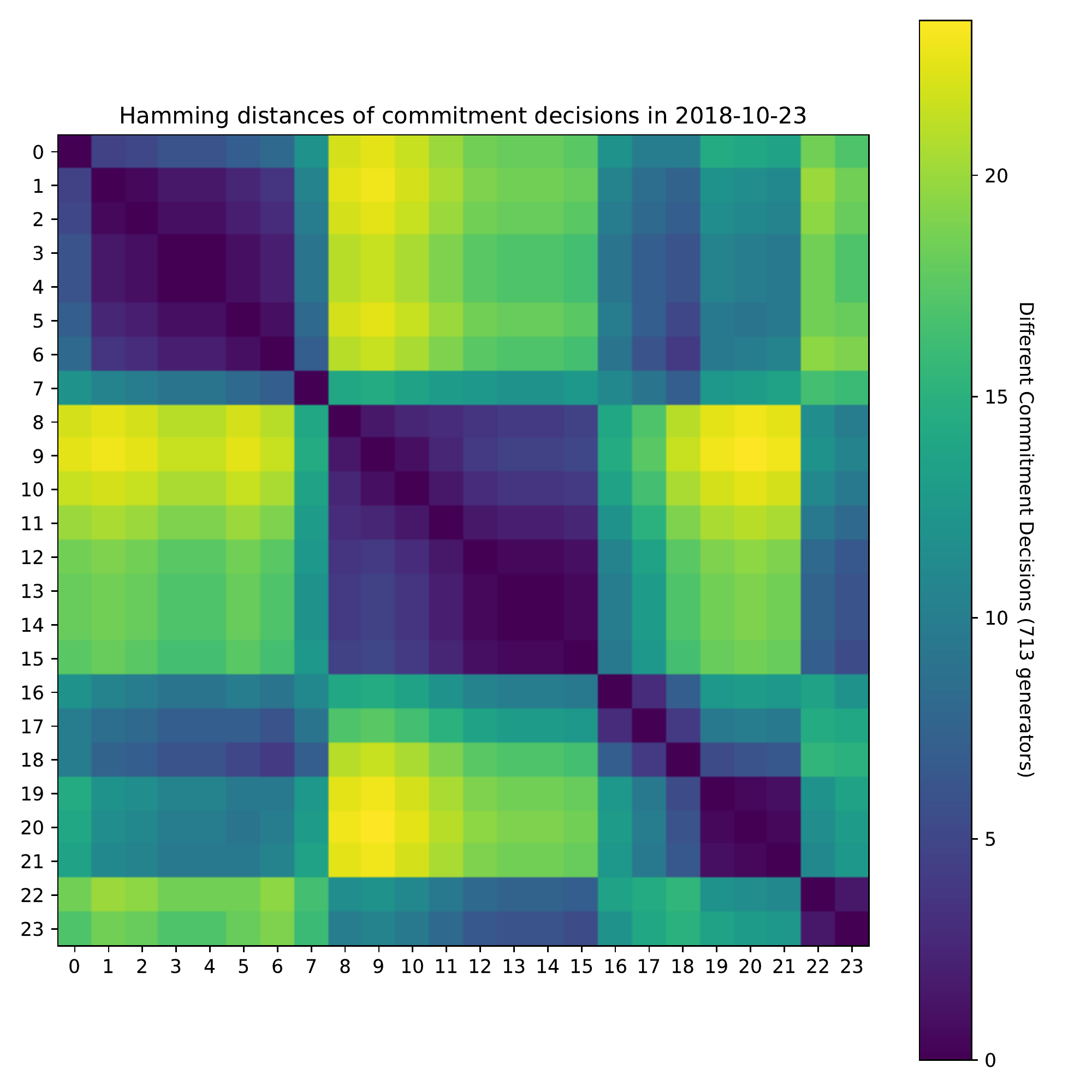}
\caption{Heat map of commitment decisions across hours}
\label{fig:heat_map_commitment_decisions}
\end{figure*}

\begin{figure*}[!h]
\centering
\includegraphics[width=0.30\linewidth]{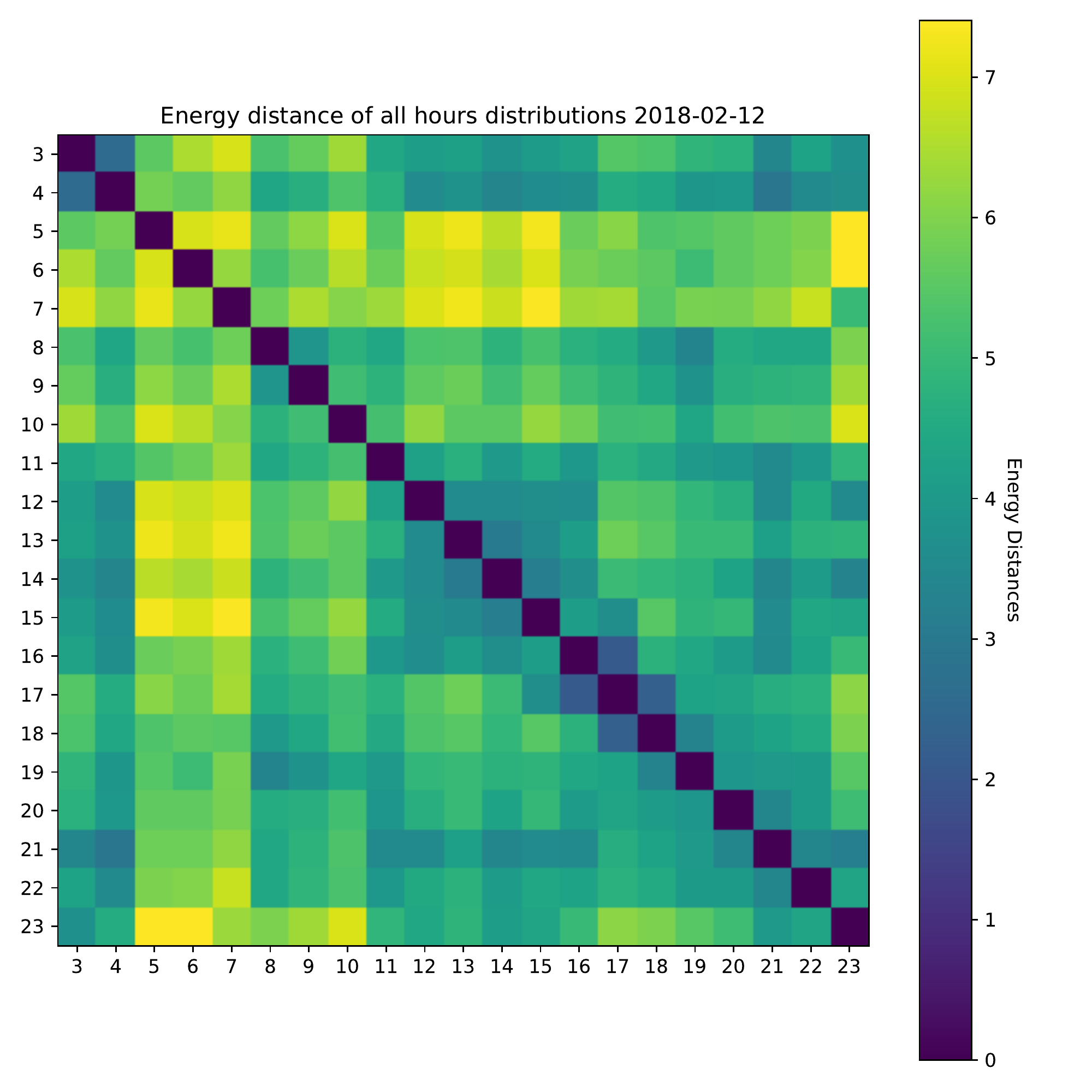}
\includegraphics[width=0.30\linewidth]{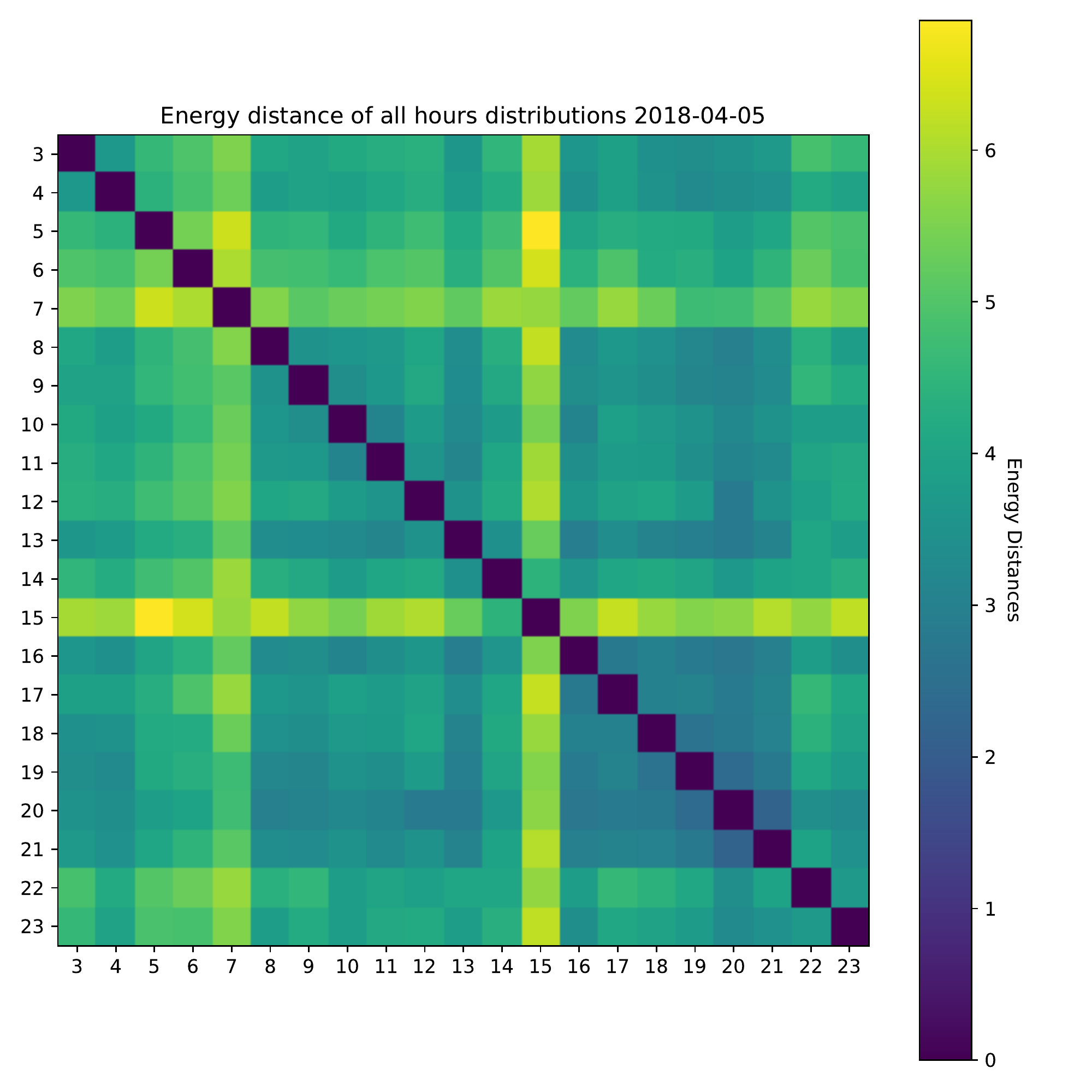} \\
\includegraphics[width=0.30\linewidth]{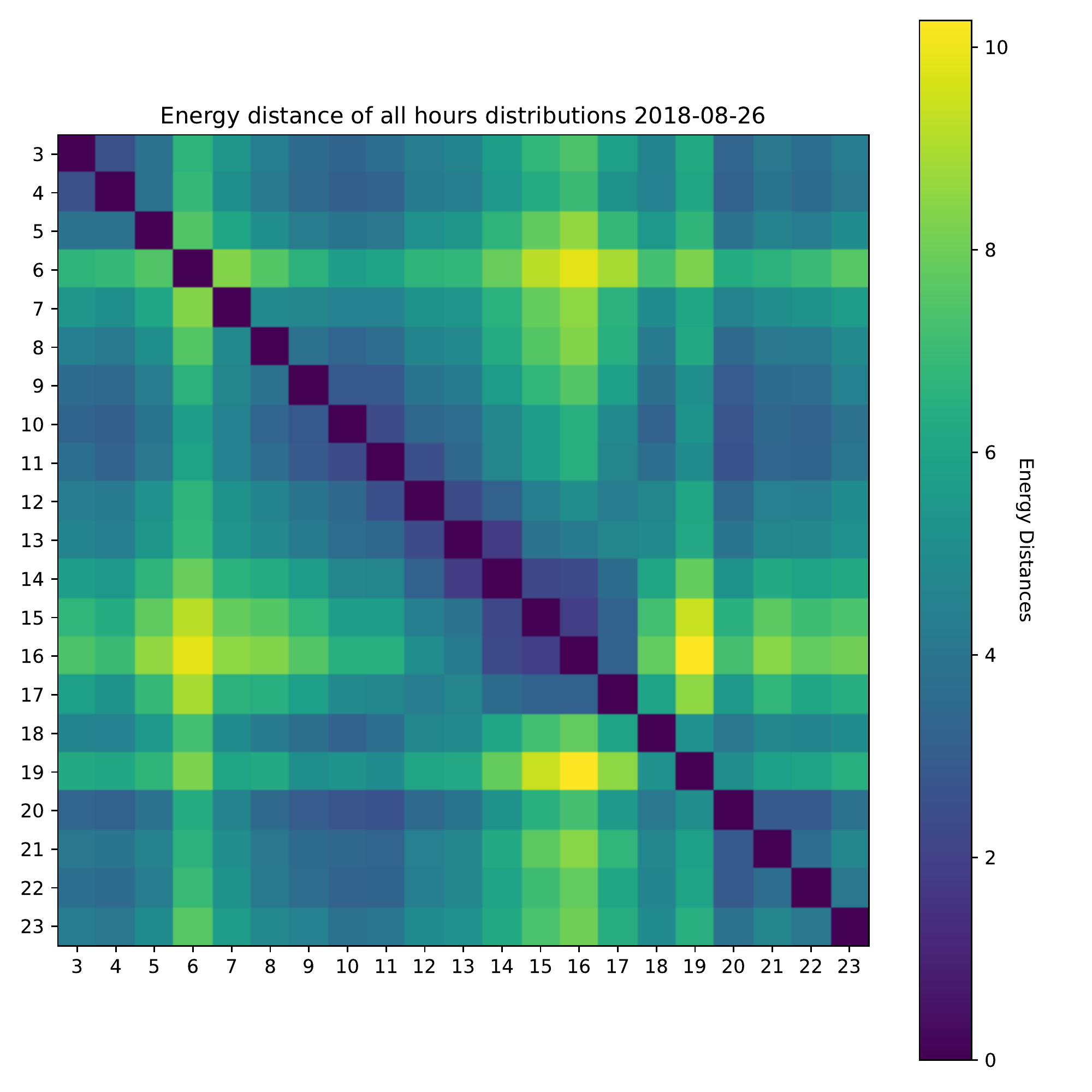}
\includegraphics[width=0.30\linewidth]{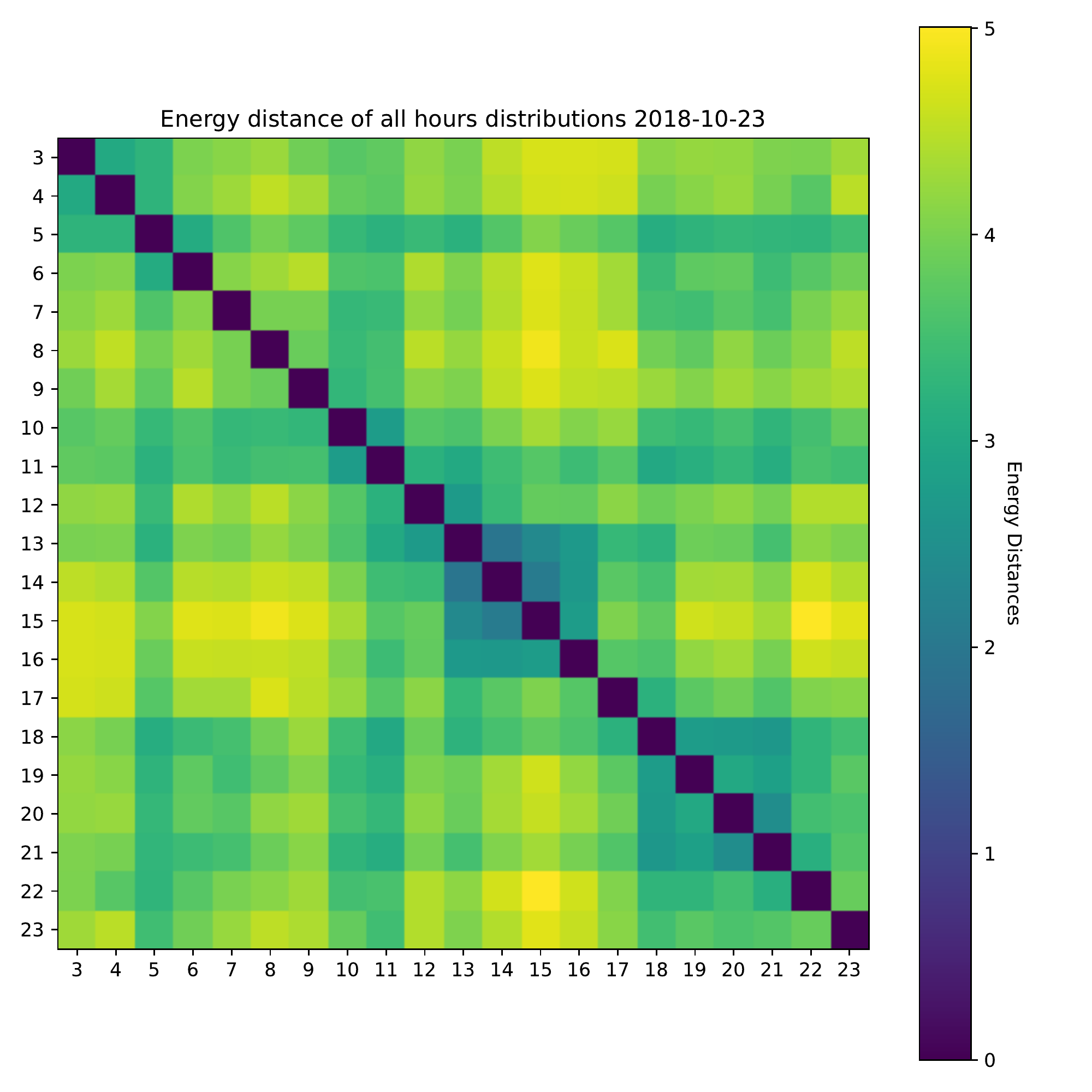}
\caption{Heat map of the energy distances of hourly distribution across hours}
\label{fig:heat_map_energy_distance}
\end{figure*}

The distribution shifts between different hours are measured using the hamming distance of commitment decisions and the energy distance of the input features, shown in Figure~\ref{fig:heat_map_commitment_decisions} and \ref{fig:heat_map_energy_distance}, respectively. Each entry indicates the distance between the corresponding hours. Therefore, the diagonal entries are always 0 since the distances between the same dataset are $0$. Interestingly, the block-wise patterns can be observed for energy distances, which is more obvious for hamming distances. Within the block which consists of several consecutive hours, the distribution shift tends to be small. Cross the block, the distribution shift largely. This, to some degree, explains why the CTR models scale well when the span of the combined dataset is less than certain hours, from the data distribution perspective.

\end{document}